\documentclass[letterpaper]{article} 
\usepackage{aaai2026}  
\usepackage{times}  
\usepackage{helvet}  
\usepackage{courier}  
\usepackage[hyphens]{url}  
\usepackage{amssymb}
\usepackage{graphicx} 
\usepackage{natbib}  
\usepackage{caption} 
\usepackage{amsmath}
\usepackage{amsthm}
\newtheorem{definition}{Definition}
\usepackage{algorithm}
\usepackage{algpseudocode}
\algrenewcommand\algorithmicrequire{\textbf{Input:}}
\algrenewcommand\algorithmicensure{\textbf{Output:}}

\usepackage{cleveref}
\nocopyright

\usepackage{booktabs}
\usepackage{multirow}
\usepackage{graphicx}
\usepackage{cleveref}
\usepackage{tikz}
\usepackage{bm}
\usepackage{makecell}

\usepackage{newfloat}
\usepackage{listings}
\DeclareCaptionStyle{ruled}{labelfont=normalfont,labelsep=colon,strut=off} 
\floatstyle{ruled}
\newfloat{listing}{tb}{lst}{}
\floatname{listing}{Listing}

\title{TriShGAN: Enhancing Sparsity and Robustness in Multivariate Time Series Counterfactuals Explanation}

\author{Hongnan Ma, Yiwei Shi, Guanxiong Sun, Mengyue Yang, Weiru Liu\\
}
\affiliations{
    University of Bristol
}

\usepackage{bibentry}

\begin{document}

\maketitle

\begin{abstract}
In decision-making processes, stakeholders often rely on counterfactual explanations, which provide suggestions about what should be changed in the queried instance to alter the outcome of an AI system. However, generating these explanations for multivariate time series (MTS) presents challenges due to their complex, multi-dimensional nature. Traditional Nearest Unlike Neighbor (NUN)-based methods typically substitute subsequences in a queried time series with influential subsequences from an NUN, which is not always realistic in real-world scenarios due to the rigid direct substitution. Counterfactual with Residual Generative Adversarial Networks (CounteRGAN)-based methods aim to address this by learning from the distribution of observed data to generate synthetic counterfactual explanations.  However, these methods primarily focus on minimizing the cost from the queried time series to the counterfactual explanations and often neglect the importance of distancing the counterfactual explanation from the decision boundary. This oversight can result in explanations that no longer qualify as counterfactual if minor changes occur within the model. To generate a more robust counterfactual explanation, we introduce TriShGAN, under the CounteRGAN framework enhanced by the incorporation of triplet loss. This unsupervised learning approach uses distance metric learning to encourage the counterfactual explanations not only to remain close to the queried time series but also to capture the feature distribution of the instance with the desired outcome, thereby achieving a better balance between minimal cost and robustness. Additionally, we integrate a Shapelet Extractor that strategically selects the most discriminative parts of the high-dimensional queried time series to enhance the sparsity of counterfactual explanation and efficiency of the training process. Our experiments across four datasets demonstrate that the introduction of a Shapelet Extractor and triplet regularization improves both the sparsity and robustness of the generated explanations. 
\end{abstract}


\section{Introduction}
Explainable Artificial Intelligence (XAI) has gained increasing popularity for enhancing the transparency and trustworthiness of black-box models. Counterfactual explanation (CFE) \cite{shi2025counterfactual} provides actionable insights by identifying \textit{how} modifications to the queried instance can lead to a desired outcome of AI system \cite{aggarwal2010inverse,guidotti2024counterfactual}. While numerous CFE methods have been proposed for models trained on tabular and image data, the area of time series data, particularly MTS, has been largely overlooked due to its complex nonlinear temporal dependencies and multi-dimensional nature \cite{saluja2021towards}. In real-world applications such as medical diagnostics and stock forecasting, understanding model decisions is crucial for ensuring trust and accountability in AI systems \cite{krollner2010financial, awad2020predicting, di2023explainable,shi2025autonomous}. 

Existing CFE methods for MTS primarily fall into two categories: NUN-based methods and CounteRGAN-based methods. \textbf{NUN-based methods} focus on selecting the NUN with the target class as a CF candidate, and then replacing the most influential subsequence of the NUN with the corresponding region in the queried time series $X_{\text{queried}}$ to generate a closely aligned CFE \cite{delaney2021instance}. Meanwhile, \textbf{CounteRGAN-based methods} apply a RGAN in conjunction with a pre-trained classifier. Instead of generating complete synthetic CFEs, RGAN learns to produce residuals that represent minimal yet meaningful perturbations. These residuals are then added back to $X_{\text{queried}}$ to generate the $X_{\text{cf}}$. RGAN generates CFEs by optimizing a minimax objective between a generator and a discriminator, enabling the model to approximate complex probability distributions through backpropagation and gradient descent \cite{nemirovsky2022countergan}. 

\begin{figure}[htbp]
\centering
\includegraphics[width=7cm]{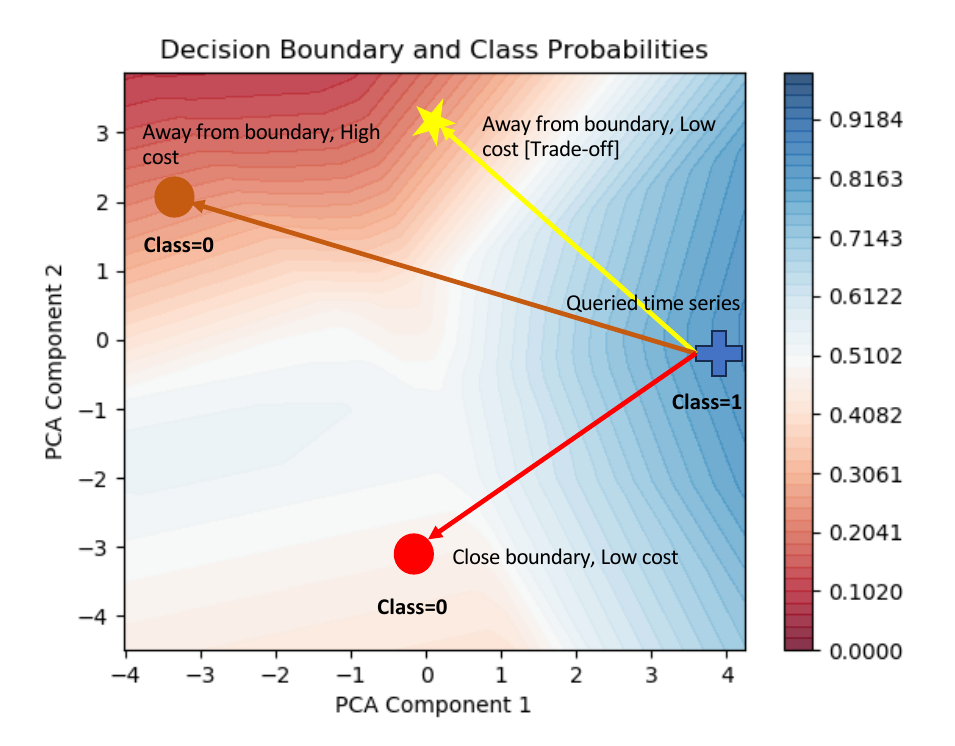}
\caption{The yellow, brown, and red points represent three viable counterfactuals for the $X_{\text{queried}}$, indicated by the blue cross point. Each counterfactual satisfies the condition of differing predicted classes from the original. Among these, the yellow pentagram is identified as the optimal choice due to its greater distance from the decision boundary and higher similarity to $X_{\text{queried}}$. Conversely, the brown point, despite sharing the same class probability with the yellow pentagram, but with high cost. The red cross point, while close to the $X_{{\text{queried}}}$, is positioned near the decision boundary, making it sensitive to minor variations and less desirable for stable counterfactual analysis.}

\label{boundary}
\end{figure}

Despite their effectiveness in generating CFE, these methods face significant limitations. CFE generated by NUN-based methods often enforces rigid constraints, such as substituting a single time series \cite{ates2021counterfactual} or a fixed-length subsequence of all signals \cite{li2023attention}, limiting their flexibility. It is not always practical in real-world scenarios. For instance, in medical diagnostics, it is both impractical and unethical to substitute time series subsequences from one patient with those of another. While CounteRGAN-based methods solve this issue by learning from the distribution of observed data, this approach aims at minimizing cost relative to the $X_{\text{queried}}$ and tends to overlook the importance of distancing the generated CFEs from decision boundaries and ensuring robustness. The decision boundary, a conceptual surface that delineates data points across different class labels, is visualized as a white band between classes as shown in Figure \ref{boundary}. The proximity of CFEs to these boundaries indicates a high sensitivity to minor modifications in model parameters or during retraining which lead to these CFEs invalid, significantly limiting their reliability \cite{dutta2022robust}. Although increasing the cost associated with generating a CFE can enhance its robustness in linear models, the trade-offs between cost and robustness in complex neural networks still need to be explored \cite{dutta2022robust}.

To overcome the constraints of traditional NUN-based methods, we introduce a CounteRGAN-based framework that synthesizes CFE with improved robustness and efficiency. By adopting the triplet loss method from unsupervised learning, we map its three elements—anchor, positive samples, and negative samples—to the counterfactual domain. Specifically, we treat $X_{\text{queried}}$ as the anchor, which is optimized as $X_{\text{cf}}$ at the conclusion of the process; factual examples serve as positive samples, while counterfactual examples act as negative samples. During training, the model minimizes the distance between $X_{\text{cf}}$ and factual samples while maximizing the distance from counterfactual samples. This strategy based on distance metric ensures the $X_{\text{cf}}$ is close to $X_{\text{queried}}$ and forces the model to learn feature distributions that are critical for achieving the counterfactual outcome, thereby balancing the trade-off between cost and robustness of the generated CFEs. Furthermore, to address the challenge of redundancy in high-dimensional MTS, which often leads the model to modify uninformative regions, we implement a pre-training phase using a Shapelet Extractor to extract highly discriminative time series subsequences. It allows the model to strategically pinpoint and extract class-specific subsequences. This targeted approach not only augments the sparsity but also boosts the effectiveness.

\begin{figure*}[htbp]
\centering
\makebox[\textwidth]{\includegraphics[width=19cm]{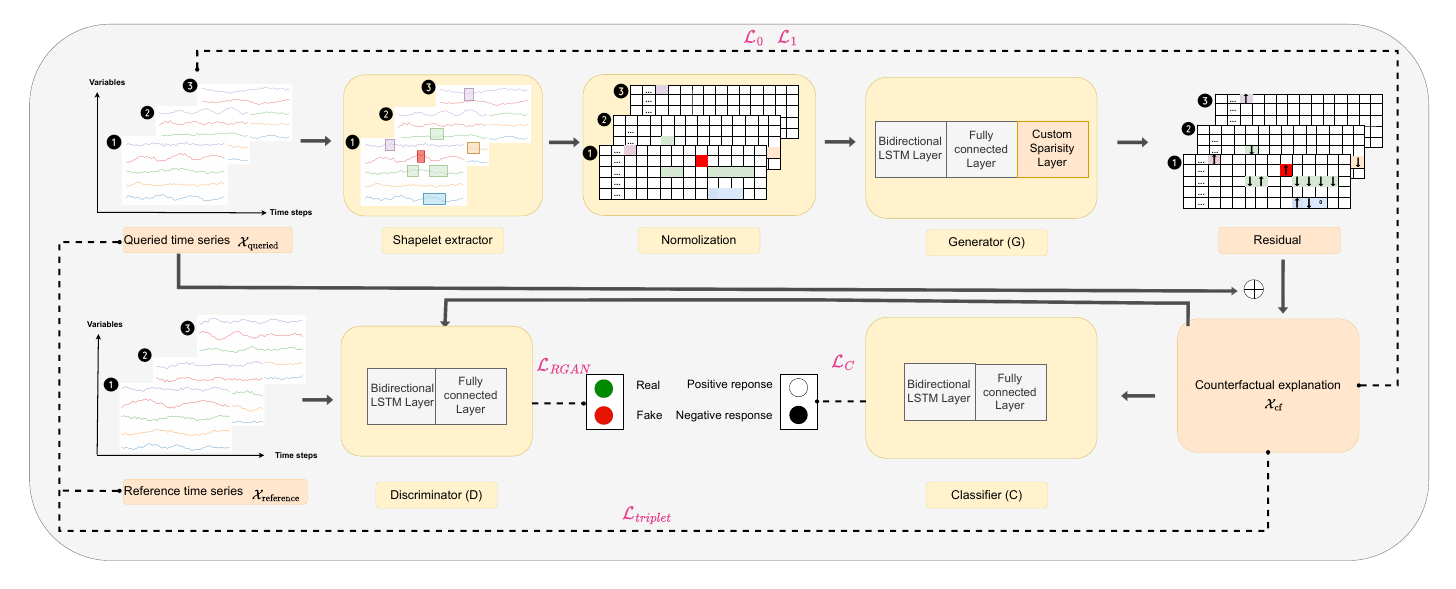}}
\vspace{-10mm}
\caption{The architecture of TriShGAN.}
\label{framework}
\end{figure*}

The contributions of this paper are threefold: First, we identify and discuss the limitations of traditional NUN-based and CounteRGAN-based methods for generating CFEs. Second, as illustrated in Figure \ref{framework}, we introduce a novel framework - \textbf{TriShGAN} (built on CounteR\textbf{GAN}, incorporates \textbf{Tri}plet loss within the generator, along with a pre-trained \textbf{Sh}apelet extractor), to generate sparse and robust CFEs for MTS more efficiently. Third, we validate our approach through comprehensive experiments on four real-world datasets, comparing our method against existing baselines. The results demonstrate that our framework not only addresses previous limitations but also efficiently balances the trade-off between cost and robustness.

\section{Related Work}
\label{related_work}
We present extended analyses of the two additional CFE methods for MTS - the iterative optimization explanation and the NUN‑based explanation in Appendix A. In this section, we concentrate on the method most closely related to our approach.

\subsection{CounteRGAN-based Explanation}
GANs, widely used for generating realistic synthetic images, train a generator to create fake samples that a discriminator mistakenly classifies as real \cite{goodfellow2020generative}. However, when applied to generating CFE for MTS, traditional GANs often suffer from mode collapse, where the generator fails to explore the entire data distribution, leading to a lack of diversity in the generated counterfactuals. The CounteRGAN \cite{nemirovsky2022countergan} introduces an RGAN, where the generator learns to produce residuals rather than complete instances. This approach aligns closely with the concept of minimal perturbations in counterfactual search, helping to mitigate mode collapse. Additionally, CounteRGAN integrates a pre-trained target classifier, ensuring that generated counterfactuals successfully flip the model’s prediction. However, applying CounteRGAN directly to MTS counterfactual generation can lead to low sparsity because of using symmetric activation functions in the generator model. To our knowledge, \cite{lang2023generating} is the only work that builds on CounteRGAN to make it more suitable for MTS CFEs. They propose a framework called SPARSE, which customizes the sparsity mechanism in the generator by introducing a sparsity layer with two interacting ReLU activations. Instead of a single linear output, their dual ReLU configuration allows the network to produce both positive and negative residuals, enabling fine-grained regularization directly on the residuals rather than on the entire counterfactual instance. However, the trade-off between metrics was not previously considered. We address this issue by incorporating triplet loss and a pre-trained Shapelet Extractor.

\section{Preliminaries}
\label{problem_form}

\subsection{Multivariate Time Series}

\begin{definition}[Multivariate Time Series]\label{def:MTS}
    A multivariate time series (MTS) $X = \{\mathbf{x}_1, \dots, \mathbf{x}_V\} \in \mathbb{R}^{T \times V}$ consists of $V$ univariate time series signals, each of length $T$. Here, $\mathbf{x}_i = \{x_{i,1}, \dots, x_{i,T}\} \in \mathbb{R}^T$ denotes the $i$-th univariate time series signal.
\end{definition}

\begin{definition}[Time Series Classification Dataset]
    A time series classification dataset $\mathcal{D} = (\mathcal{X}, \mathbf{y})$ consists of a set of $n$ MTS $\mathcal{X} = \{X_1, \dots, X_n\} \in \mathbb{R}^{n \times V \times T}$, where each MTS has $V$ signals and $T$ time steps. The labels $\mathbf{y} = \{y_1, \dots, y_n\} \in \{1, \dots, c\}^n$ are the corresponding class labels for the $n$ MTS, with $c$ being the number of classes. We have binary classification when $c = 2$ and multi-class classification when $c > 2$. For each MTS $X_n$, the value of the $v$-th signal at the $t$-th time step is denoted as $x_{v,t}^{(n)}$.
\end{definition}

\subsection{Subsequence and Distance} 

\begin{definition}[Time Series Subsequence]\label{def:sub}
Based on \Cref{def:MTS}, given a $i$-th unvarite time series signal $\mathbf{x}_i = \{x_{i,1}, \dots, x_{i,T}\}$ of length \( T \), a consecutive subsequence $\mathbf{x}_i[j : j+l-1] = \{x_{i,j}, \dots, x_{i,j+l-1}\}$ is a segment of the time series $\mathbf{x}_i$, where \(x_{i,j} \) is the start index, \(x_{i,j+l-1} \) is the end index, and \( 1 \leq j \leq j+l-1 \leq T \).
\end{definition}

\begin{definition}[Minimum subsequence distance (MSD)]
Let $\mathbf{x}_{i}=(x_{i1},\ldots,x_{iT})$ be a time series of length $T$
and let $\mathbf{s}$ be a query subsequence of length $l\le T$.
The MSD between $\mathbf{s}$ and $\mathbf{x}_{i}$ is defined as
$$
\text{MSD}(\mathbf{x}_{i},\mathbf{s})
    = \min_{1\le j\le T-l+1}
      \text{CID} \!\bigl(\mathbf{x}_{i}[j:j+l-1],\,\mathbf{s}\bigr),
$$
where $\text{CID} (\cdot,\cdot)$ denotes the complexity‑invariant distance (See Appendix B for details).

\end{definition}

\subsection{Shapelet}

\begin{definition}[Shapelet Candidate]
A shapelet candidate \( S_{ij} \) is a time series subsequence extracted from a multivariate time series \( X \). It is defined as:
\begin{align}
    S_{ij} = \mathbf{x}_i[j : j + l - 1] = \{ x_{i,j}, \dots, x_{i,j+l-1} \}
\end{align}
A shapelet candidate is typically extracted using \textit{Perceptually Important Points (PIPs)} which was proposed by ~\cite{le2024shapeformer} (See Appendix B for details). 
\end{definition}

\begin{definition}[Shapelet]
A shapelet is a shapelet candidate \( S_{ij} \) that provides maximum discriminative power for a classification task. The discriminative ability of a shapelet is evaluated based on its \textit{Information Gain (IG)}, computed using an optimal split point \( \text{OSP}(S_{ij}) \) \cite{hills2014classification}.   Formally, a shapelet \( S_{ij} \) satisfies:
\begin{align}
    \text{IG}(S_{ij}, \text{OSP}(S_{ij})) \geq \text{IG}(S_{k\ell}, \text{OSP}(S_{k\ell})), \quad \forall k \neq i \text{ or } \ell \neq j.
\end{align}
where \( S_{k\ell} \) represents any other shapelet candidate. Details of the $\operatorname{OSP}$ computation are provided in Appendix B.
\end{definition}

\begin{definition}[Shapelet Pool]
    
In a time series classification task with \( c \) classes and the training set \( \mathcal{D}_{\text{train}} = \langle \mathcal{X}_{\text{train}}, \mathbf{y}_{\text{train}} \rangle \), the shapelet pool \( \mathcal{S} \) is the collection of the most discriminative shapelets across all classes:
\begin{align}
    \mathcal{S} = \bigcup_{p=1}^{c} \mathcal{S}_p,
\end{align}
where \( \mathcal{S}_p \) is the set of top \( g \) shapelets selected from shapelet candidates \( S_{ij} \) for class \( p \), based on achieving the highest \textit{IG}:
\begin{align}
    \mathcal{S}_p = \{ \tilde{S}_p^1, \tilde{S}_p^2, \dots, \tilde{S}_p^g \}
\end{align}
Each \( \tilde{S}_p^k \) is a selected shapelet that maximizes class separability for class \( p \).
\end{definition}

\begin{definition}[Shapelet Extractor]
Given a queried time series $X_{\text{queried}} \in \mathbb{R}^{V \times T}$ with its label $y$ from $\mathcal{D}_{\text{test}} = \langle \mathcal{X}_{test}, \mathbf{y}_{test} \rangle$, the Shapelet Extractor calculates MSD between $X_{\text{queried}}$ and each of the \( g \) shapelets in the set \( \mathcal{S}_y \). For each shapelet $\tilde{S}_y^k$, in \( \mathcal{S}_y \), the discriminative subsequence in $X_{\text{queried}}$ is extracted based on MSD. A set of discriminative subsequence of $X_{\text{queried}}$ is defined as \( I(X_{\text{queried}}) = \{S_y^1, S_y^2, \dots, S_y^g\} \). 
\end{definition}

\section{TriShGAN}
\label{methodology}

\subsection{TriShGAN Architecture}
\label{architecture}
The architecture of TRiShGAN is illustrated in Figure \ref{framework} and the pseudocode is showed in Appendix C\footnote{The source code will be made public upon the paper’s acceptance}. The pipeline contains three steps.

\textbf{Shapelet Extraction and Normalization.} Initially, before training the generator $G$ and discriminator $D$, a classifier $C$ and a Shapelet Extractor are pre-trained. The training dataset is divided into $\mathcal{X}_{\text{queried}}$  and $\mathcal{X}_{\text{reference}}$ based on their labels. For instance, time series in $\mathcal{X}_{\text{queried}}$ are labelled as 1, and those in $\mathcal{X}_{\text{reference}}$ are labelled as 0. $\mathcal{X}_{\text{queried}}$ goes through the Shapelet Extractor to isolate a set of discriminative subsequences $I(X_{\text{queried}})$ for each queried time series in $\mathcal{X}_{\text{queried}}$, represented visually by rectangles (second left box on top in Figure \ref{framework} ). Subsequences outside these rectangles are normalized to zero (middle box on top is depicted as white grids). The discriminative subsequences are then concatenated with zeros, padding the base until reaching the original time series data's length. 

\textbf{Counterfactual Generation.} The reformatted $\mathcal{X}_{\text{queried}}$ is subsequently fed into the $G$ (second right box on top). $G$ is based on the bidirectional LSTM neural network, following the approach described by \cite{lang2023generating}. They added a custom sparsity layer with dual ReLU activation in the generator to generate more sparse CFE. $G$ computes residuals $G(\mathcal{X}_{\text{queried}})$ for each query, where $G(\mathcal{X}_{\text{queried}})$ signifies the magnitude of change required in each time series in $\mathcal{X}_{\text{queried}}$. Positive residuals are represented by $\uparrow$, negative by $\downarrow$, and no change by zero (first right box on top). These residuals are added back to $\mathcal{X}_{\text{queried}}$ to generate the counterfactual instances $\mathcal{X}_{\text{cf}}$ (first right box on the bottom).

\textbf{Counterfactual Evaluation.} Once the counterfactual set $\mathcal{X}_{\text{cf}}$ is generated, it is used in two ways. First, it is assessed by classifier $C$ to verify whether the label has flipped (positive response indicates that the label has been flipped successfully). Secondly, the $\mathcal{X}_{\text{cf}}$ and $\mathcal{X}_{\text{reference}}$ are evaluated in the discriminator $D$, which strives to differentiate the generated counterfactuals from actual reference sequences, thus enhancing the realism of the CFs (Green point represents that the generated synthetic CFs are realistic enough to deceive the discriminator). The $D$ is designed as the bidirectional LSTM neural network as well. To further optimize the generated CFs, several regularization terms are incorporated into the training process: $\mathcal{L}_{RGAN}$, $\mathcal{L}_{C}$, $\mathcal{L}_0$, $\mathcal{L}_1$, and $\mathcal{L}_{\text{triplet}}$.

\subsection{Counterfactual Triplet Loss}
\begin{figure}[htbp]
\centering
\includegraphics[width=0.4\textwidth]{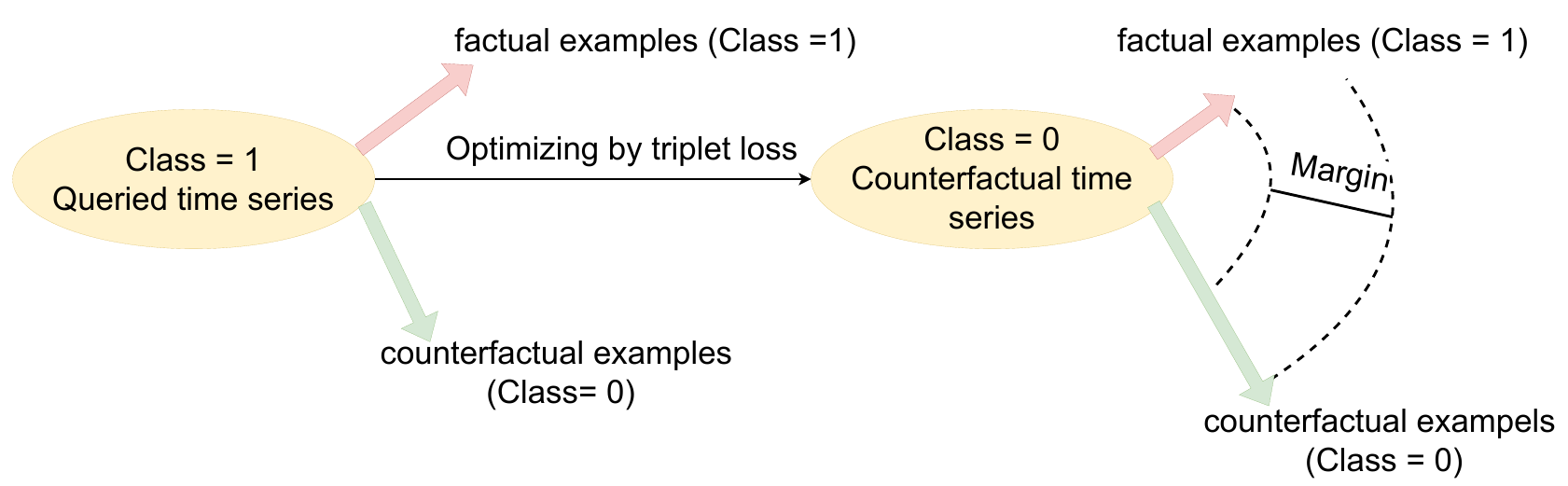}
\caption{Illustration of the process of optimizing triplet loss} 
\label{triplet_loss}
\end{figure}

Triplet loss, initially introduced by \cite{weinberger2009distance}, has been adapted to learn embeddings that effectively bring similar data points closer in the embedding space while ensuring that dissimilar points remain well-separated. 
We employ triplet loss build on \cite{liu2024explaining, li2021shapenet} to generate robust CFE for MTS. Figure \ref{triplet_loss} illustrates the process of optimizing triplet loss. We define a triplet in CF domain which considers the distances between three types of examples: an anchor example (anchor starts as $X_{\text{queried}}$ and is optimized by $\phi$ through training to become the $X_{\text{cf}}$), positive examples (factual examples), and negative examples (counterfactual examples). 

\begin{definition}[Queried and reference time series]  
A dataset $\mathcal{D}_{\text{train}}$ can be divided into two subsets as follows:  
\begin{align}
\mathcal{D_{\text{train}}} = (\mathcal{X}_{\text{queried}}, \mathbf{y}_{\text{queried}}) \cup (\mathcal{X}_{\text{reference}},\mathbf{y}_{\text{reference}}),
\end{align}
where  $\mathbf{y}_{\text{queried}} \neq \mathbf{y}_{\text{reference}}$.

\end{definition}

\begin{definition}[Factual examples]
   Given a pre-trained classifier \( C \) and a queried time series \( X_i \), let \( \mathcal{X}_i^{=} \) be a set of instances. A subset of instances \( \{X_1^{=}, X_2^{=}, \ldots, X_n^{=}\} \subset \mathcal{X}_i^{=} \) are defined as \textit{factual examples} if each \( X_i^{=} \) satisfies \( C(X_i^{=}) = C(X_i) \) and they are the \( n \) \textbf{closest} instances to \( X_i \) based on $\mathcal{L}_2$ distance. 

\end{definition}

\begin{definition}[Counterfactual examples]
    Given a pre-trained classifier \( C \) and a queried time series \( X_i \), let \( \mathcal{X}_i^{\neq} \) be a set of instances. A subset of instances \( \{X_1^{\neq}, X_2^{\neq}, \ldots, X_n^{\neq}\} \subset \mathcal{X}_i^{\neq} \) are defined as \textit{ counterfactual examples} if each \( X_i^{\neq} \) satisfies \( C(X_i^{\neq}) \neq C(X_i) \) and they are \( n \) instances \textbf{randomly} selected from the set of all instances that meet this criterion. 
\end{definition}

During training, each batch contains factual and counterfactual examples sampled from $\mathcal{X}_{\text{queried}}$ and $\mathcal{X}_{\text{reference}}$, respectively. We set the ratio of number of factual and counterfacutal sampels to be equal. The number of elements in a triplet, denoted as $n$, should not exceed the batch size $N$. Under this setup, each triplet \( t_i \) is formulated as:
\begin{align}  
t_i = \{\phi(X_i), \left\{ \mathcal{X}_i^{\neq} \right\}, \left\{\mathcal{X}_i^{\neq} \right\}\}.
\end{align}
Accordingly, the triplet loss can be defined as:
\begin{align}
\mathcal{L}_{\text{triplet}} = \max\left(0, d\left(\phi(X_i), \mathcal{X}_i^{=}\right) - d\left(\phi(X_i), \mathcal{X}_i^{\neq}\right) + \gamma\right),
\end{align}
where \( d \) calculates the Manhattan distance. The distance from the anchor to the factual samples \( \mathcal{X}_i^= \) is computed as $d(\phi(X_i), \mathcal{X}_i^=) = \frac{1}{n} \sum_{i=1}^{n} | \phi(X_i) - {X_i^=} |$. Similarly, the distance from the anchor to the counterfactual samples \( \mathcal{X}_i^\neq \) is: $d(\phi(X_i), \mathcal{X}_i^\neq) = \frac{1}{n} \sum_{i=1}^{n} | \phi(X_i) - {X_i^\neq} |$. \( \gamma \) is a predefined margin that specifies the minimum difference required between the distances of different-class and same-class pairs. Optimizing the triplet loss ensures that the distance between the anchor and negative examples is significantly greater than the distance to positive examples. This encourages the model to capture meaningful feature distributions of the instance with the desired outcome, rather than merely minimizing cost. These learned patterns drive the CFE deeper into the target class region, enhancing robustness and preventing borderline classifications.

\subsection{Final loss}

The objective function of CounteRGAN is composed of several terms, tailored to guide the training process more effectively:
\begin{equation}
\label{loss}
    \mathcal{L}_{CGAN}(G,D) = \mathcal{L}_{RGAN}(G,D) + \mathcal{L}_{C}(G,C) + \mathcal{L}_{Reg}(G),
\end{equation}
where $G$, $D$, and $C$ denote the generator, discriminator, and classifier networks, respectively. $\mathcal{L}_{RGAN}(G,D)$ addresses the traditional adversarial loss, $\mathcal{L}_C(G,C)$ targets the classification accuracy, and $\mathcal{L}_{Reg}(G)$ applies regularization to promote desirable properties in the generated samples. Following the protocols in CounteRGAN \cite{nemirovsky2022countergan}, we leverage residual GAN loss and classifier loss to alleviate mode collapse and ensure the generated time series achieves the desired class transformation. Specifically, the residual GAN loss is formulated as:
\begin{equation}
\begin{aligned}
\mathcal{L}_{RGAN}(G, D) = & \, \mathbb{E}_{X \sim p_{\text{data}}}[\log D(X)] \\
& + \mathbb{E}_{X \sim p_{\text{data}}}[ \log (1 - D(X + G(X)))],
\end{aligned}
\end{equation}
where $X$ represents the queried time series, and $G(X)$ is the residual added to $X$ to generate the counterfactual. The generator and discriminator are trained simultaneously in an adversarial minimax fashion. 

The classifier loss is calculated as:
\begin{align}
\nonumber\mathcal{L}_{C}(G, C) &= \mathbb{E}_{X \sim p_{\text{data}}} \left[-y \log(C(X + G(X))) \right. \\ & \left. - (1-y) \log(1 - C(X + G(X))) \right],
\end{align}
where $C(\cdot)$ denotes the classifier's output, and $y$ is the desired class label for the counterfactuals. This loss helps steer the generated modifications towards fulfilling the desired outcome effectively.

The final loss formulation is formulated as:
\begin{align}
    \mathcal{L}_{\text{TriShGAN}}(G,D) =\ & \lambda_1  \mathcal{L}_{\text{triplet}} + \lambda_2 \mathcal{L}_{\text{RGAN}}(G,D) \nonumber \\
    & + \lambda_3 \mathcal{L}_C(G,C) + \lambda_4 \mathcal{L}_0 + \lambda_5 \mathcal{L}_1,
\end{align}
where $\mathcal{L}_0$ and $\mathcal{L}_1$ denote $\mathcal{L}_0$ norm and Manhattan distance. $\mathcal{L}_0$ is used to generate sparse CFEs as sparsity loss. $\mathcal{L}_1$ is used as similarity loss. Additional weighting factors $\lambda_1$ to $\lambda_5$ allow for the individual components of the generator loss to be adjusted or toggled on and off, catering to the specific needs of individual datasets. 


\section{Experiments Setup}
\label{setup}
All experimental results for our methods, baselines, and ablation studies are reported as the mean $\pm$ standard deviation from 5 repetitions. For each metric in the results, we use $\uparrow$ to indicate a preference for higher values and $\downarrow$ for lower values. The best results are marked in \textbf{bold}. This framework uses the Adam optimizer with a learning rate of $1 \times 10^{-5}$ and a batch size of 64. We set all $\lambda$ equal to 1 across all GAN-based baselines and datasets for consistency. This allows us to see if the framework naturally balances robustness and proximity without relying on carefully tuning or over-weighting certain terms. 


\subsection{Dataset}
The evaluation was conducted on four publicly available MTS datasets from the UEA MTS archive\footnote{\url{https://www.timeseriesclassification.com/}}: FingerMovements (FM), HeartBeat (HB), SelfRegulationSCP1 (SCP1), and SelfRegulationSCP2 (SCP2). These datasets span diverse domains and are structured for binary classification tasks, making them ideal for assessing the effectiveness of our approach across varied applications. Details, including their number of dimensions, time series length, and the sizes of the training and test splits, are provided in Appendix E. Additionally, we follow the number of shapelets and window sizes that achieved the highest accuracy ranking as found in \cite{le2024shapeformer}. 


\subsection{Baseline} 
We evaluate our method against other baseline methods, including a NUN-based method AB-CF \cite{li2023attention}, and some GAN-based explanation methods such as GAN \cite{goodfellow2020generative}, CounteRGAN \cite{nemirovsky2022countergan} and SPARSE \cite{lang2023generating}, using several metrics (See Appendix D for details).

\subsection{Evaluation Metric}
For each $X_\text{queried}$ in the test dataset, every method generates two outputs: the counterfactual explanation $X_{\text{cf}}$ and its associated prediction probability $P_{\text{cf}}$. We utilize $X_{\text{cf}}$ to assess the quality of the CFEs produced by different methods, comparing them directly with $X_\text{queried}$. Meanwhile, $P_{\text{cf}}$ is used to quantify the robustness of the generated counterfactuals.

\begin{itemize}
    \item \textbf{Target Class Validity (TCV):} TCV quantifies the percentage of counterfactual instances that successfully alter the original prediction of the model. A higher TCV indicates better effectiveness in changing model decisions. \cite{mahajan2019preserving, guo2023counternet}

    \item \textbf{Proximity:} This metric measures the closeness between the $X_\text{queried}$ and the $X_\text{cf}$, computed using the $\mathcal{L}_1$ norm. It assesses the minimal changes required to alter the prediction, normalized by the number of time steps and signals in the dataset. Lower values suggest higher proximity, indicating subtle modifications.

    \item \textbf{Sparsity:} Sparsity is evaluated based on the number of modified time steps and signals required to transform a $X_\text{queried}$ into its $X_\text{cf}$, by using the $\mathcal{L}_0$ norm. Lower values indicate higher sparsity.

    \item \textbf{Local Outlier Factor (LOF)}: LOF quantifies how much a $X_\text{cf}$ deviates from the norm within a baseline dataset \(\mathcal{X}_{\text{test}}\). The LOF greater than 1 indicates that \(X_\text{cf}\) is less dense than its neighbors, classifying it as an outlier \cite{patcha2007overview}.

    \item \textbf{Robustness:} We quantify the robustness of $X_\text{cf}$ using \( P_{\text{cf}} \). In our experiment, the label of $X_\text{queried}$ is 1, and the desired label is 0. The probability \( P_{\text{cf}} \) is obtained from the model’s output after applying the sigmoid function. Therefore, the decision boundary is set at 0.5. If \( P_{\text{cf}} \) is distant from the 0.5, it indicates that the CFE is robust. Specifically, the lower the value of \( P_{\text{cf}} \), the more robust \cite{jiang2024robust} (See Appendix D for details on LOF and Robustness).
\end{itemize}

\section{Experiment Result}
\label{result}

\subsection{Performance of TriShGAN}

\subsubsection{TCV}

Table \ref{exp_results} compares the TCV achieved by TriShGAN and various baseline methods across four datasets. AB-CF, GAN, and our method all achieve 100\% TCV on each dataset. AB-CF is a NUN-based explanation method. For this type of method, the worst-case scenario involves considering the NUN as the CFE, which explains why the TCV is consistently 100\%. With GAN, pattern collapse presents a significant issue where the generator focuses on deceiving the discriminator rather than capturing the full diversity of the data distribution, often at the expense of other metrics (Proximity and Sparsity). Our approach outperforms both CounterGAN and SPARSE by ensuring that the effectiveness in generating diverse counterfactuals does not compromise other metrics.

\subsubsection{Trade off between robustness and proximity}
\begin{figure}[htbp]
\centering
\includegraphics[scale=0.28]{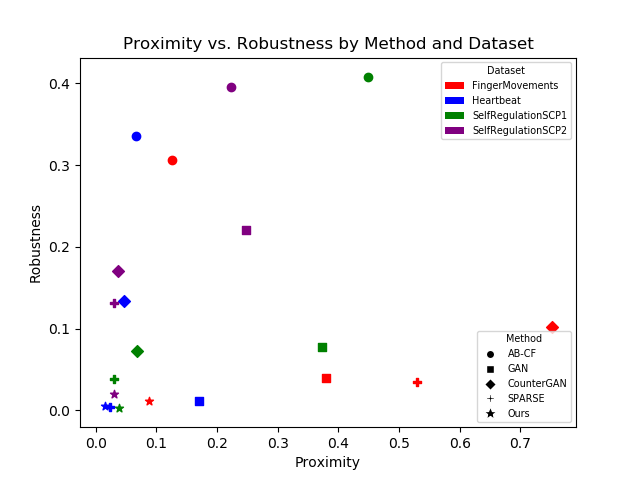}
\caption{Trade off between proximity and robustness} 
\label{tradeoff}
\end{figure}

Figure \ref{tradeoff} illustrates the trade-off between proximity and robustness across various methods and datasets. This visualization distinctly positions our method in the lower-left quadrant consistently across all datasets, indicating its good performance in achieving both minimal proximity and maximal robustness. In comparison, the SPARSE method exhibits limitations in maintaining robustness in certain datasets such as SelfRegulationSCP2 and FingerMovements. On the other hand, the AB-CF method represented by circles in the upper region, tends to prioritize proximity while often neglecting the robustness of the generated counterfactuals. The performance of GAN and CounteRGAN methods exhibits variability, failing to consistently secure a favorable trade-off across all datasets. These observations underscore the effectiveness of our method's integration of triplet loss during the training process, enabling it to adeptly balance the cost-robustness trade-off. 

\subsubsection{Sparsity and plausibility}
\begin{figure}[htbp]
\centering
\begin{minipage}{.25\textwidth}
  \centering
  \includegraphics[scale=0.19]{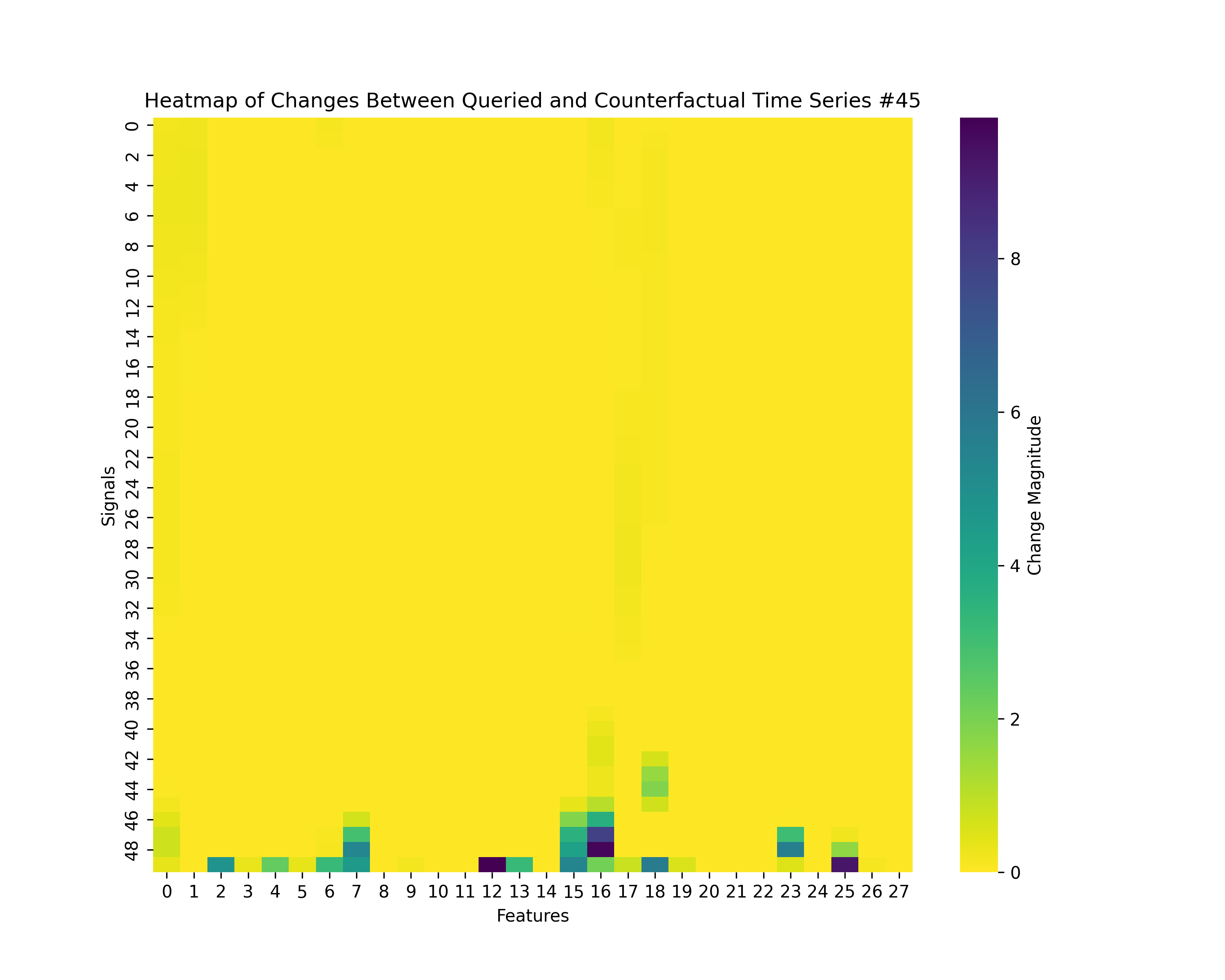}
  \caption{FM - TriShGAN}
  \label{fig:finger_our}
\end{minipage}%
\begin{minipage}{.25\textwidth}
  \centering
  \includegraphics[scale=0.19]{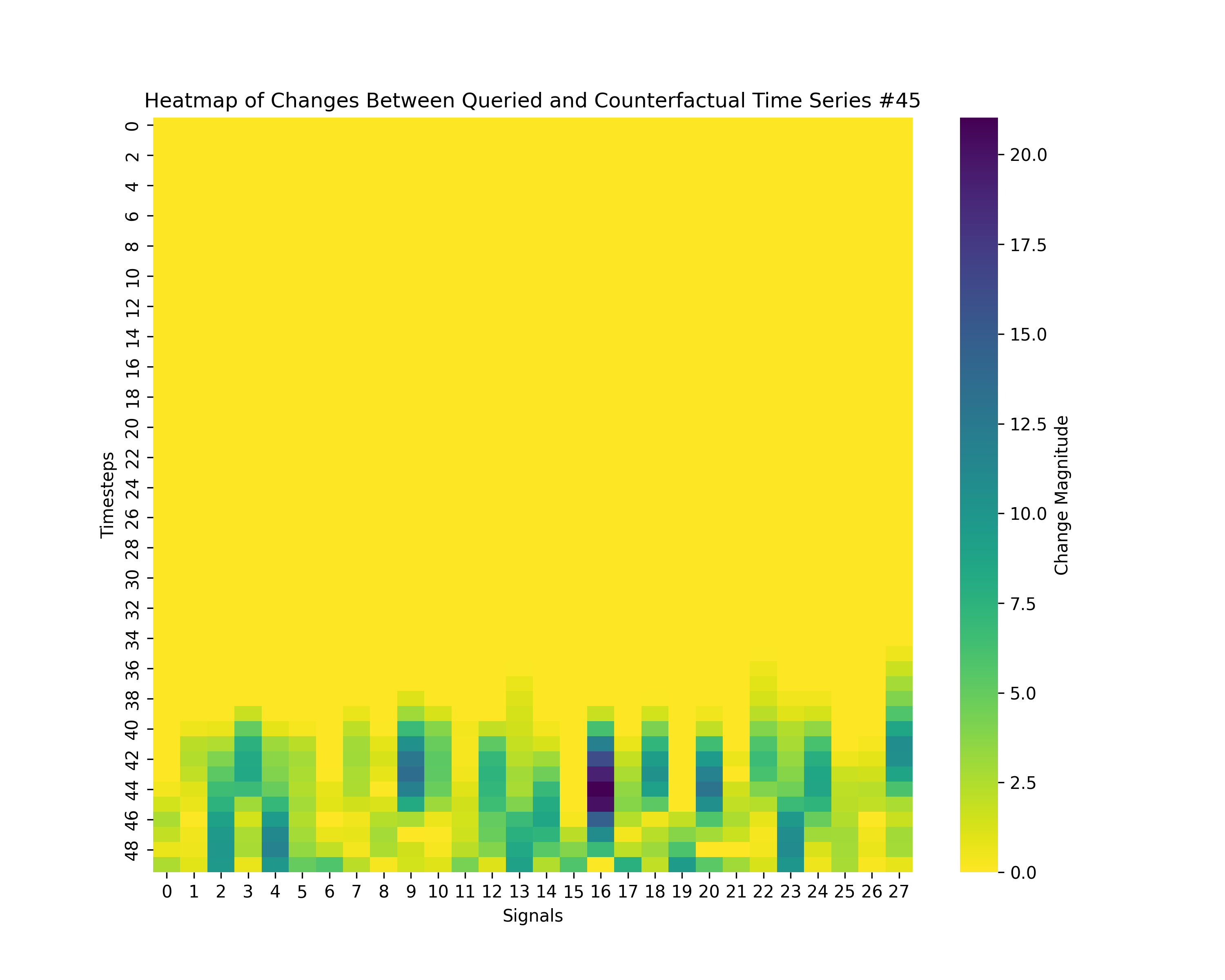}
  \caption{FM - SPARSE}
  \label{fig:finger_sparse}
\end{minipage}
\end{figure}

To show the details of how $X_{\text{queried}}$ changed to $X_{\text{cf}}$, we utilized a queried time series from the FM dataset to demonstrate the details. The variations in color represent the magnitude of residual changes: The ligher color indicates smaller changes. If the colors appear less dense across the heatmap, this indicates CFE is more sparse. SPARSE in Figure \ref{fig:finger_sparse} captured the necessary modifications in the later subsequence of the time series to generate $X_{\text{cf}}$. However, TriShGAN showed in Figure \ref{fig:finger_our} distinctly focuses on significant signals rather than all signals. In contrast, SPARSE, lacking initial discriminative subsequence constraints, applies transformations across all signals. For the plausibility metric, all methods exhibit good performance; details are analysed in Appendix F.

\begin{table}[htbp]
\centering
\caption{Experiment Results Across Datasets and Methods}
\label{exp_results}
\small
\scriptsize                 
\setlength{\tabcolsep}{0.5pt}
\begin{tabular}{lcccccc} 
\toprule
\multirow{2}{*}{\makecell[c]{Data\\Set}} & \multirow{2}{*}{Method} & \multicolumn{5}{c}{Metrics} \\ 
\cmidrule(l){3-7}
 & & TCV $\uparrow$ & Robustness $\downarrow$ & Proximity $\downarrow$ & Sparsity $\downarrow$ & Plausibility $\downarrow$\\
\midrule
\multirow{5}{*}{FM} 
 & AB-CF  & \(100_{\,\pm0.00}\) & \(0.31_{\,\pm0.04}\) & \(0.13_{\,\pm0.03}\) & \(0.46_{\,\pm0.10}\) & \(0.05_{\,\pm0.03}\) \\
 & GAN    & \(100_{\,\pm0.91}\) & \(0.04_{\,\pm0.01}\) & \(0.38_{\,\pm0.01}\) & \(1.00_{\,\pm0.00}\) & \(1.00_{\,\pm0.00}\) \\
 & C-GAN  & \(98_{\,\pm2.04}\)  & \(0.10_{\,\pm0.03}\) & \(0.75_{\,\pm0.55}\) & \(1.00_{\,\pm0.00}\) & \(1.00_{\,\pm0.00}\) \\
 & SPARSE & \(100_{\,\pm0.00}\) & \(0.04_{\,\pm0.01}\) & \(0.53_{\,\pm0.03}\) & \(0.11_{\,\pm0.00}\) & \(0.40_{\,\pm0.01}\) \\
 & Ours   & \(\bm{100_{\,\pm0.00}}\) & \(\bm{0.01_{\,\pm0.01}}\) & \(\bm{0.09_{\,\pm0.02}}\) & \(\bm{0.04_{\,\pm0.01}}\) & \(\bm{0.04_{\,\pm0.00}}\) \\
\midrule
\multirow{5}{*}{HB} 
 & AB-CF  & \(100_{\,\pm0.00}\) & \(0.34_{\,\pm0.01}\) & \(0.07_{\,\pm0.01}\) & \(0.37_{\,\pm0.07}\) & \(\bm{0.00_{\,\pm0.00}}\) \\
 & GAN    & \(100_{\,\pm0.00}\) & \(0.01_{\,\pm0.00}\) & \(0.17_{\,\pm0.00}\) & \(1.00_{\,\pm0.00}\) & \(0.00_{\,\pm0.00}\) \\
 & C-GAN  & \(100_{\,\pm0.00}\) & \(0.13_{\,\pm0.03}\) & \(0.05_{\,\pm0.01}\) & \(1.00_{\,\pm0.00}\) & \(0.05_{\,\pm0.03}\) \\
 & SPARSE & \(100_{\,\pm0.00}\) & \(\bm{0.004_{\,\pm0.00}}\) & \(0.023_{\,\pm0.006}\) & \(0.01_{\,\pm0.00}\) & \(0.42_{\,\pm0.00}\) \\
 & Ours   & \(\bm{100_{\,\pm0.00}}\) & \(0.006_{\,\pm0.003}\) & \(\bm{0.016_{\,\pm0.003}}\) & \(\bm{0.006_{\,\pm0.00}}\) & \(0.51_{\,\pm0.00}\) \\
\midrule
\multirow{5}{*}{SCP1} 
 & AB-CF  & \(100_{\,\pm0.00}\) & \(0.41_{\,\pm0.04}\) & \(0.45_{\,\pm0.14}\) & \(0.72_{\,\pm0.18}\) & \(0.12_{\,\pm0.13}\) \\
 & GAN    & \(100_{\,\pm0.00}\) & \(0.08_{\,\pm0.01}\) & \(0.37_{\,\pm0.00}\) & \(1.00_{\,\pm0.00}\) & \(\bm{0.00_{\,\pm0.00}}\) \\
 & C-GAN  & \(98.4_{\,\pm2.20}\) & \(0.07_{\,\pm0.02}\) & \(0.07_{\,\pm0.02}\) & \(1.00_{\,\pm0.00}\) & \(0.05_{\,\pm0.03}\) \\
 & SPARSE & \(98.6_{\,\pm1.18}\) & \(0.04_{\,\pm0.00}\) & \(\bm{0.030_{\,\pm0.01}}\) & \(\bm{0.009_{\,\pm0.00}}\) & \(0.01_{\,\pm0.00}\) \\
 & Ours   & \(\bm{100_{\,\pm0.00}}\) & \(\bm{0.003_{\,\pm0.001}}\) & \(0.038_{\,\pm0.008}\) & \(0.012_{\,\pm0.002}\) & \(0.007_{\,\pm0.00}\) \\
\midrule
\multirow{5}{*}{SCP2} 
 & AB-CF  & \(100_{\,\pm0.00}\) & \(0.40_{\,\pm0.03}\) & \(0.22_{\,\pm0.07}\) & \(0.44_{\,\pm0.13}\) & \(0.08_{\,\pm0.09}\) \\
 & GAN    & \(100_{\,\pm0.00}\) & \(0.22_{\,\pm0.10}\) & \(0.25_{\,\pm0.01}\) & \(1.00_{\,\pm0.00}\) & \(\bm{0.00_{\,\pm0.00}}\) \\
 & C-GAN  & \(100_{\,\pm0.00}\) & \(0.17_{\,\pm0.01}\) & \(0.04_{\,\pm0.01}\) & \(1.00_{\,\pm0.00}\) & \(0.00_{\,\pm0.00}\) \\
 & SPARSE & \(100_{\,\pm0.00}\) & \(0.13_{\,\pm0.06}\) & \(\bm{0.030_{\,\pm0.01}}\) & \(0.008_{\,\pm0.01}\) & \(0.01_{\,\pm0.00}\) \\
 & Ours   & \(\bm{100_{\,\pm0.00}}\) & \(\bm{0.02_{\,\pm0.01}}\) & \(0.031_{\,\pm0.00}\) & \(\bm{0.008_{\,\pm0.00}}\) & \(0.01_{\,\pm0.00}\) \\
\bottomrule
\end{tabular}
\end{table}

\subsection{Result Analysis of Shapelet Extractor}
\label{6.2}
\subsubsection{Without Shapelet Extractor}
We conducted an ablation study on the Shapelet Extractor, and the results are presented in Table \ref{ablation_shapelet}. The methods with the Shapelet Extractor achieved a 100\% TCV across all datasets, indicating complete success in flipping labels for every dataset. Furthermore, methods that utilized shapelets consistently demonstrated lower probabilistic values across all datasets, signifying that the generated counterfactuals are more robust, representing a 50\% improvement compared to methods without the Shapelet Extractor. Additionally, the sparsity values of the framework with the Shapelet Extractor were superior to those without it across almost all datasets, except for SelfRegulaSCP1, where the value was slightly lower but still reasonable. Moreover, our approach, which perturbs only the subsequence that has the most significant impact on classification, minimizes disturbances to the overall data distribution and reduces the likelihood of generating out-of-distribution samples.

\begin{table}[t]
\centering
\caption{Ablation Experiment for Shapelet Extractor
         \newline\small (Within each dataset, the first row is \textit{with} and the second row is \textit{w/o}.)}

\label{ablation_shapelet}
\scriptsize               
\setlength{\tabcolsep}{0.5pt}   
\begin{tabular}{lccccc}
\toprule
\makecell[c]{Data\\Set} & TCV $\uparrow$ & Robustness $\downarrow$ &
Proximity $\downarrow$ & Sparsity $\downarrow$ & Plausibility $\downarrow$\\
\midrule
FM   & \(100_{\,\pm0.00}\) & \(0.04_{\,\pm0.01}\) & \(\bm{0.08_{\,\pm0.01}}\) & \(0.11_{\,\pm0.00}\) & \(0.40_{\,\pm0.01}\)\\
     & \(100_{\,\pm0.00}\) & \(\bm{0.02_{\,\pm0.00}}\) & \(0.09_{\,\pm0.02}\) & \(\bm{0.05_{\,\pm0.01}}\) & \(\bm{0.05_{\,\pm0.01}}\)\\
\midrule
HB   & \(100_{\,\pm0.00}\) & \(0.004_{\,\pm0.001}\) & \(0.023_{\,\pm0.006}\) & \(0.010_{\,\pm0.003}\) & \(0.42_{\,\pm0.00}\)\\
     & \(100_{\,\pm0.00}\) & \(\bm{0.003_{\,\pm0.001}}\) & \(\bm{0.016_{\,\pm0.004}}\) & \(\bm{0.006_{\,\pm0.002}}\) & \(0.42_{\,\pm0.00}\)\\
\midrule
SCP1 & \(98_{\,\pm1.18}\)  & \(0.04_{\,\pm0.00}\) & \(\bm{0.03_{\,\pm0.01}}\) & \(\bm{0.009_{\,\pm0.004}}\) & \(0.009_{\,\pm0.004}\)\\
     & \(\bm{100_{\,\pm0.00}}\) & \(\bm{0.02_{\,\pm0.01}}\) & \(0.04_{\,\pm0.02}\) & \(0.010_{\,\pm0.006}\) & \(\bm{0.007_{\,\pm0.000}}\)\\
\midrule
SCP2 & \(100_{\,\pm0.00}\) & \(0.13_{\,\pm0.06}\) & \(\bm{0.030_{\,\pm0.011}}\) & \(0.008_{\,\pm0.011}\) & \(0.011_{\,\pm0.00}\)\\
     & \(100_{\,\pm0.00}\) & \(\bm{0.067_{\,\pm0.052}}\) & \(0.034_{\,\pm0.016}\)  & \(\bm{0.006_{\,\pm0.003}}\) & \(0.011_{\,\pm0.00}\)\\
\bottomrule
\end{tabular}
\end{table}

\begin{table}[b]
\centering
\caption{Ablation Experiment for Triplet Loss}
\label{ablation_triplet}
\scriptsize                 
\setlength{\tabcolsep}{0.5pt}
\begin{tabular}{lccccc}
\toprule
\makecell[c]{Data\\Set} & TCV $\uparrow$ & Robustness $\downarrow$ & Proximity $\downarrow$ & Sparsity $\downarrow$ & Plausibility $\downarrow$\\
\midrule
FM & \(54.43_{\,\pm5.36}\)  & \(0.21_{\,\pm0.02}\)  & \(\bm{0.09_{\,\pm0.01}}\) & \(0.038_{\,\pm0.004}\) & \(0.21_{\,\pm0.02}\)\\
   & \(\bm{97.96_{\,\pm0.00}}\) & \(\bm{0.05_{\,\pm0.03}}\) & \(0.10_{\,\pm0.03}\) & \(\bm{0.036_{\,\pm0.006}}\) & \(\bm{0.10_{\,\pm0.01}}\)\\
\midrule
HB & \(\bm{99.42_{\,\pm0.83}}\) & \(0.066_{\,\pm0.047}\) & \(\bm{0.01_{\,\pm0.00}}\) & \(0.02_{\,\pm0.02}\) & \(\bm{0.42_{\,\pm0.01}}\)\\
   & \(100_{\,\pm0.00}\)       & \(\bm{0.004_{\,\pm0.001}}\) & \(0.02_{\,\pm0.01}\) & \(\bm{0.01_{\,\pm0.00}}\) & \(0.43_{\,\pm0.01}\)\\
\midrule
SCP1 & \(26.71_{\,\pm0.00}\) & \(0.29_{\,\pm0.01}\) & \(\bm{0.001_{\,\pm0.000}}\) & \(\bm{0.001_{\,\pm0.000}}\) & \(0.03_{\,\pm0.00}\)\\
  & \(\bm{100_{\,\pm0.00}}\) & \(\bm{0.01_{\,\pm0.01}}\) & \(0.019_{\,\pm0.004}\) & \(0.011_{\,\pm0.005}\) & \(\bm{0.01_{\,\pm0.00}}\)\\
\midrule
SCP2  & \(44.82_{\,\pm2.31}\) & \(0.44_{\,\pm0.01}\) & \(0.018_{\,\pm0.001}\) & \(\bm{0.003_{\,\pm0.001}}\) & \(\bm{0.02_{\,\pm0.02}}\)\\
  & \(\bm{100_{\,\pm0.00}}\) & \(\bm{0.01_{\,\pm0.01}}\) & \(\bm{0.003_{\,\pm0.001}}\) & \(0.011_{\,\pm0.000}\) & \(0.19_{\,\pm0.02}\)\\
\bottomrule
\end{tabular}
\end{table}


\subsection{Result Analysis of Triplet Loss}
\label{6.3}
\subsubsection{Without triplet loss}
To rigorously assess whether triplet loss contributes to generating robust counterfactuals, we deactivated the classifier's loss function in our experiments. If our approach still demonstrates strong performance under these conditions—without updates to the classifier's parameters—it suggests that the ability of triplet loss to identify distinctive feature distributions from the desired class does not heavily rely on the classifier. This implies that our method is robust across various classifiers. Table \ref{ablation_triplet} shows that if triplet loss was omitted, the results were unsatisfactory, with probabilistic consistently exceeding 0.2. Conversely, in experiments that incorporated triplet loss, even without the influence of classifier loss, the generated counterfactuals were effectively kept away from the decision boundary, underscoring the effectiveness of triplet loss in enhancing the robustness of the generated explanations.

\subsubsection{Parameter sensitive analysis for triplet loss}
There are two hyperparameters in triplet loss - $n$ and $\gamma$. We define the central margin as
\begin{equation}
\gamma_\text{central} = \frac{1}{2} \left| \frac{1}{N} \sum_{i=1}^N d(\phi(X_i), \mathcal{X}_i^{\neq}) - \frac{1}{N} \sum_{i=1}^N d(\phi(X_i), \mathcal{X}_i^{=}) \right|.
\end{equation}

In our experiments, we set $\Delta = 10^{\left\lfloor \log_{10}\gamma_{\text{central}} \right\rfloor}$, and construct the candidate margin set: $$S(\gamma_{\text{central}}) = \bigl\{
        \gamma_{\text{central}} - \Delta,\,
        \gamma_{\text{central}},\,
        \gamma_{\text{central}} + \Delta,\,
        \gamma_{\text{central}} + 2\Delta
      \bigr\}.$$

\begin{figure}[t]
\centering
\includegraphics[scale=0.14]{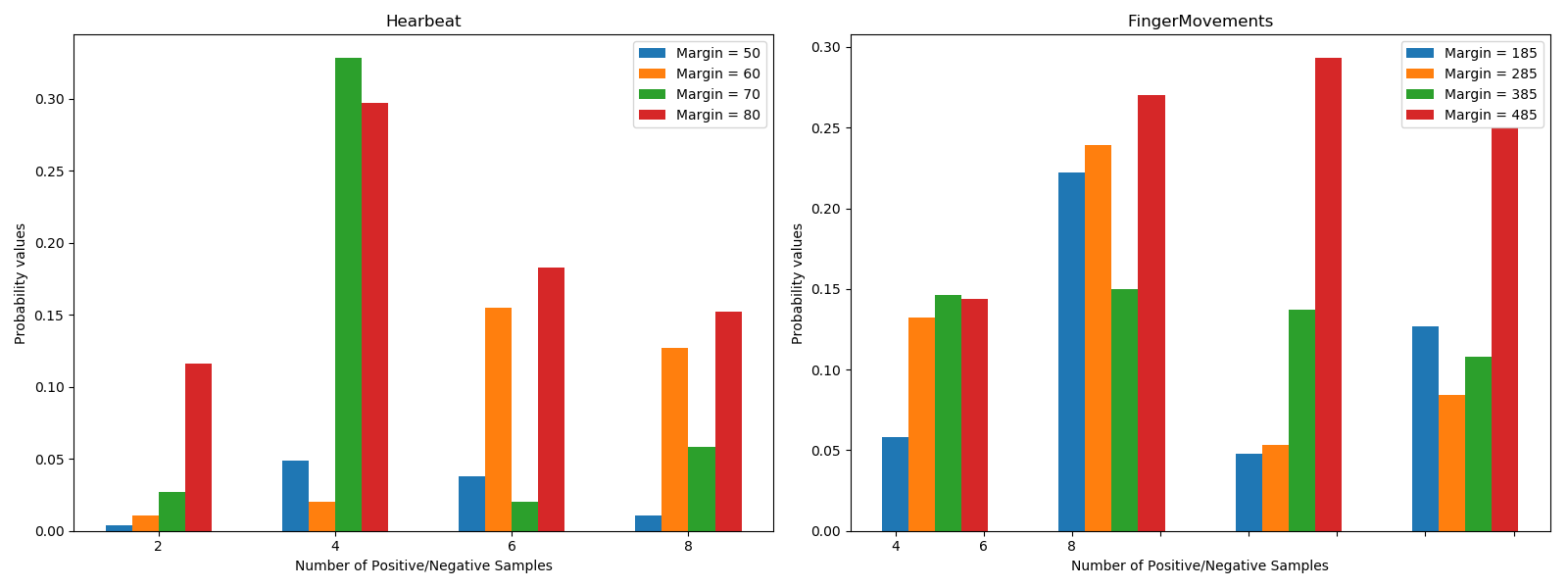}
\caption{Probability change between different $n$ and $\gamma$} 
\label{prob}
\end{figure}

We choose the $n$ as ${2,4,6,8}$ and show the results of HB and FM, shown in Figure \ref{prob}, which indicates that a lower $n$ tends to generate more robust CFEs, as it also reduces the noise introduced by the samples. Having multiple negatives can lead to noise during the feature distribution learning process, preventing the model from precisely learning the distinguishing characteristics of different categories, ultimately resulting in less robust CFEs. Furthermore, we found that margins slightly below the $\gamma_{\text{central}}$ performed relatively better. Excessively large margins decrease robustness because they may force the model to emphasize differences in features that are not key to distinguishing between two categories, leading to less robust counterfactuals.

\section{Conclusion}
\label{conclusion}
We propose TriShGAN for generating more sparse and robust MTS CFEs. Our approach maps triplet loss within the counterfactual domain, enabling the effective capture of distinguishing features across classes and producing more robust outcomes. Additionally, considering the complexity of high-dimensional MTS, we incorporated a Shapelet Extractor to swiftly identify and focus on discriminative subsequences during training. Extensive experiments show our method outperforms existing approaches in balancing multiple metrics. Future extensions could include providing textual counterfactual explanations and expand the scope beyond merely focusing on binary classification.

\newpage
\makeatletter
\@ifundefined{isChecklistMainFile}{
  \newif\ifreproStandalone
  \reproStandalonetrue
}{
  \newif\ifreproStandalone
  \reproStandalonefalse
}
\makeatother

\ifreproStandalone
\setlength{\pdfpagewidth}{8.5in}
\setlength{\pdfpageheight}{11in}
\frenchspacing

\fi
\setlength{\leftmargini}{20pt}
\makeatletter\def\@listi{\leftmargin\leftmargini \topsep .5em \parsep .5em \itemsep .5em}
\def\@listii{\leftmargin\leftmarginii \labelwidth\leftmarginii \advance\labelwidth-\labelsep \topsep .4em \parsep .4em \itemsep .4em}
\def\@listiii{\leftmargin\leftmarginiii \labelwidth\leftmarginiii \advance\labelwidth-\labelsep \topsep .4em \parsep .4em \itemsep .4em}\makeatother

\setcounter{secnumdepth}{0}
\renewcommand\thesubsection{\arabic{subsection}}
\renewcommand\labelenumi{\thesubsection.\arabic{enumi}}

\newcounter{checksubsection}
\newcounter{checkitem}[checksubsection]

\newcommand{\checksubsection}[1]{%
  \refstepcounter{checksubsection}%
  \paragraph{\arabic{checksubsection}. #1}%
  \setcounter{checkitem}{0}%
}

\newcommand{\checkitem}{%
  \refstepcounter{checkitem}%
  \item[\arabic{checksubsection}.\arabic{checkitem}.]%
}
\newcommand{\question}[2]{\normalcolor\checkitem #1 #2 \color{blue}}
\newcommand{\ifyespoints}[1]{\makebox[0pt][l]{\hspace{-15pt}\normalcolor #1}}

\bibliography{aaai2026}
\newpage
\onecolumn

\section*{Appendix}

\subsection*{Appendix Table of Contents}
\renewcommand{\arraystretch}{1.1}
\begin{tabular}{@{}p{1.2em}p{0.9\linewidth}@{}}
\textbf{A.} & Related Work \\
\textbf{B.} & Supplementary Formulas for Shapelet Extraction\\
\textbf{C.} & Pseudocode \\
\textbf{D.} & Baseline and Metrics\\
\textbf{E.} & Dataset \\
\textbf{F.} & Complete Experiment Result \\
\textbf{G.} & Compute Resources and Environment Details \\
\textbf{H.} & Further Analysis of Shapelet Extractor\\
\textbf{I.} & Hyperparameter Selection for Triplet loss\\
\end{tabular}

\section{Related Work}
We present extended analyses of the two additional CFE methods for MTS - the iterative‑optimization explanation and the NUN‑based explanation.
\subsection{Iterative Optimization Explanation}
Wachter et al. \cite{wachter2017counterfactual} were the first to propose a method for generating CFEs. Their approach minimizes a loss function consisting of two components: a prediction term, which ensures the generated counterfactual reaches the target label, and an  $\mathcal{L}_1$ distance term which maintains similarity between the CFE and the original instance. Alibi \cite{klaise2021alibi} introduced an additional autoencoder reconstruction loss, which penalizes counterfactuals that deviate significantly from the data distribution, thereby reducing the likelihood of generating outlier CFEs. While these methods can effectively explore the search space in structured tabular datasets, they become impractical in high-dimensional feature spaces due to the high computational cost associated with iterative optimization methods.

\subsection{NUN-based Explanation}
A pioneer CFE approach for time series is Native Guide \cite{delaney2021instance}, which consists of two main steps. Firstly, the NUN of the $X_{\text{queried}}$ is selected from the target class as a CF candidate. To make the CF sparser and closer to $X_{\text{queried}}$, perturbations are applied at NUN level. Specifically, the feature importance vector $w$, derived from the Class Activation Map (CAM), is used to identify the most influential contiguous subsequence within the NUN. The corresponding region in $X_{\text{queried}}$ is then modified iteratively, increasing the length of the perturbed subsequence until the label changes. Following this paradigm, some approaches replace CAM with dynamic time barycenter averaging \cite{filali2022mining} and motif-based methods \cite{li2022motif}, to find the centroid representation of the desired class that is maximally representative of that class. CoMTE \cite{ates2021counterfactual}, was originally proposed as a CFE method for MTS. It employs a heuristic search based on hill climbing, combined with a post hoc trimming step, to identify the minimal set of signals that need to be perturbed within the NUN. This approach works by disabling one signal at a time and evaluating its impact. However, it often results in unnecessary modifications, leading to low sparsity and high-cost CFEs. Recently, several shapelet-based \cite{bahri2022shapelet} and temporal rule-based \cite{bahri2022temporal} counterfactual methods have been introduced. These approaches enhance interpretability by leveraging pre-mined shapelets or temporal rules to guide modifications within the NUN. However, they often suffer from high computational costs of extracting shapelets or temporal rules. \cite{li2023attention} follows a greedy selection strategy, identifying fixed-length multivariate subsequences that maximize classification entropy when isolated and replacing them with the corresponding subsequences from the NUN in the desired class. Although these methods naturally ensure in-distribution counterfactuals since they generate CFEs by replacing parts of the original data. They suffer from a major limitation: lack of flexibility. Since these methods rigidly replace fixed subsequences in all signals of the queried time series or turn off a whole signal to obtain CFEs, they often modify unnecessary regions. As a result, their adaptability to complex real-world scenarios is significantly constrained.

\section{Supplementary Formulas for Shapelet Extraction}
\subsection{Complexity-Invariant Distance (CID)}
Following \cite{le2024shapeformer}, we use the CID as our distance metric, which is widely used in time‑series mining.
\begin{definition}[CID]
Given a time series signal $\mathbf{x}_i$, and a subsequence $\mathbf{s} = \{s_1, \dots, s_l\}$ of length $l$, with $l \leq T$, the CID between the subsequence $\mathbf{x}_i[j:j+l-1]$ and $\mathbf{s}$ is defined as the Euclidean distance between the subsequence $\mathbf{x}_i[j:j+l-1]$ and $\mathbf{s}$, scaled by a correction factor (CF) that accounts for the difference in their complexities \cite{batista2014cid}. The CID is given by:
\begin{align}
\text{CID}(\mathbf{x}_i[j:j+l-1], \mathbf{s}) = CF \times \sqrt{\sum_{q=1}^{l} (\mathbf{x}_i[j+q-1] - s_q)^2}
\end{align}
where the $CF$ is defined as:
\begin{align}
CF = \frac{\max(C(\mathbf{x}_i[j:j+l-1]), C(\mathbf{s}))}{\min(C(\mathbf{x}_i[j:j+l-1]), C(\mathbf{s}))}
\end{align}
Here, $C(\mathbf{z})$ is a measure of the complexity of the sequence $\mathbf{z}$, which can be calculated as the sum of the absolute differences between consecutive elements in $\mathbf{z}$:
\begin{align}
C(\mathbf{z}) = \sum_{q=1}^{l-1} |z_{q+1} - z_q|
\end{align}
This complexity measure adjusts the raw Euclidean distance to account for differences in the structural characteristics of the time series subsequence and the comparison subsequence $\mathbf{s}$.
\end{definition}

\subsection{Optimal Split Point (OSP)}
IG is calculated using the OSP. The OSP is a threshold distance \( d^* \) such that the separation of time series instances based on whether their \textit{Minimum Subsequence Distance (MSD)} to \( S_{ij} \) is above or below \( d^* \) maximizes class separability. The IG is calculated as:
\begin{align}
    \text{IG}(S_{ij}, d^*) = H(\mathcal{D}) - \sum_{\mathcal{D}_b \in \{ \mathcal{D}_\leq, \mathcal{D}_> \}} \frac{|\mathcal{D}_b|}{|\mathcal{D}|} H(\mathcal{D}_b),
\end{align}
where \( H(\mathcal{D}) \) is the entropy of dataset \( \mathcal{D} \), and \( \mathcal{D}_\leq \) and \( \mathcal{D}_> \) are partitions based on \( d^* \).

\subsection{Perceptually Important Points (PIPs)}
The key points known as \textit{Perceptually Important Points (PIPs)} are selected from each signal's time series \cite{le2024shapeformer}. These points include the starting and ending points of the series. The algorithm proceeds to identify additional points based on the maximum \textit{reconstruction distance}, which is defined as the perpendicular distance between a target point and a line reconstructed by the two nearest previously selected important points. These points are subsequently added to the PIPs set. Each time a new PIP is added, new shapelet candidates are extracted from all possible subsequences containing three consecutive PIPs. This addition can yield up to three new shapelet candidates for each new PIP, significantly speeding up the process. Each shapelet candidate is recorded with specific information including the signal of the shapelets, their start index, end index, and variables. During the shapelet candidates extraction phase, every shapelet candidate from the training data $\mathcal{D}_{\text{train}}$ is assessed for its discriminative power by computing the SD to all instances in the dataset. The SD is used to find the optimal information gain, and the top candidates from each class, showing the highest IG, are selected. 

\section{Pseudocode}
We provide pseudocode to clarify how the Shapelet Extractor operates and how it is incorporated with CounteRGAN-based architecture into the training process. Firstly, we adopt the class-specific transformer module from \cite{le2024shapeformer}. This approach uses a PIP-based method efficiently extract shapelets from the training set, forming a `dictionary' of key class-specific patterns. For a given input time series, these shapelets guide the search for `best-fit' subsequences via a distance metric, and the extracted subsequences are then used in the downstream classification task. Our Shapelet Extractor leverages this same `best-fit' principle to pinpoint the most discriminative segments within the input time series. Once the Shapelet Extractor is trained, we will integrate it directly into the main CounterGAN‑based architecture. Algorithm 1 outlines the one‑time offline discovery of shapelets (no trainable parameters). Afterward, this pool is used to train the Shapelet Extractor itself (Algorithm 2), which, at inference time, isolates key subsequences for any new query series. Algorithm 3 depicts the full TriShGAN training loop, in which the extractor preprocesses each input, the generator produces residual perturbations, and the discriminator and classifier supply adversarial and label‑flip objectives, respectively. Details is analyzed in main paper.

\begin{algorithm}[H]
\caption{OfflineShapeletDiscovery}
\label{alg:osd}
\begin{algorithmic}[1]
\Require Dataset $\mathcal D_{\text{train}}=\{(\mathcal{X}_i,\mathbf{y}_i)\}_{i=1}^{M}$; series length $T$; \#signals $V$; \#PIPs $k$; \#shapelets‑per‑class $g$
\Ensure  Shapelet pool $\mathcal S$

\State $C\gets\varnothing$
\ForAll{$\mathcal{X}\in\mathcal D_{\text{train}}$}                         \Comment{loop over instances}
  \For{$v=1$ \textbf{to} $V$}                   \Comment{loop over signals}
    \State $P\gets[1,T]$                        
    \For{$j=1$ \textbf{to} $k-2$}
       \State $p\gets\arg\max_{t}\text{ReconDist}(t)$ \Comment{find index $p$ from $1$ to $T$ with maximum reconstruction distance}
       \State \Call{InsertSorted}{$P,p$}
       \For{$z=0$ \textbf{to} $2$}
          \If{$1\le\textrm{idx}(p)-z$ \textbf{and} $\textrm{idx}(p)+2-z\le|P|$}
            \State $\textbf{s}\gets X[P[\textrm{idx}(p)-z]:P[\textrm{idx}(p)+2-z]]$
            \State $C\gets C\cup\{(\mathbf{s},\text{meta})\}$ \Comment{auxiliary metadata stored together with $s$}
          \EndIf
       \EndFor
    \EndFor
  \EndFor
\EndFor
\ForAll{$\mathbf{s}_i \in C$}
  \State $D \gets \bigl[\operatorname{MSD}(X, \mathbf{s}_i)\;:\; X \in \mathcal D_{\text{train}}\bigr]$    
  \State $\operatorname{InfoGain}(\mathbf{s}_i) \gets$ \Call{IG}{$D$} 
\EndFor
\ForAll{class $c$}
  \State $\mathcal S \gets \mathcal S \cup \operatorname{TopG}(C_c,\operatorname{IG})$
\EndFor 

\State \textbf{return} $\mathcal S$  
\end{algorithmic}
\end{algorithm}

\begin{algorithm}[H]
\caption{Training Shapelet Extractor}
\label{alg:shapeformer}
\begin{algorithmic}[1]

\Require Training set $\mathcal D_{\text{train}}=\{(\mathcal{X}_i,\mathbf{y}_i)\}_{i=1}^{N}$
\Ensure  Trained Shapelet Extractor parameters $\Theta$, Discriminative subsequence $U$
\State $\mathcal{S} \gets$ %
      \Call{OfflineShapeletDiscovery}{$\mathcal{D}_{\text{train}}$}

\For{$\textbf{epoch}=1$ \textbf{to} $E$}
  \For{\textbf{mini‑batch} $\mathcal B \subset \mathcal D_{\text{train}}$}
    \ForAll{$(\mathcal{X},\mathbf{y})\in\mathcal B$}
      \State $U\!\gets\Call{MSD}{X,\mathcal S}$  
      \State $Z_{\text{spe}}\!\gets\Call{ClassSpecTransformer}{U}$      
      \State $V\!\gets\Call{ConvBlocks}{X}$                           
      \State $Z_{\text{gen}}\!\gets\Call{GenericTransformer}{V}$       
      \State $z \gets \text{concat}(Z_{\text{spe}}^{\ast},Z_{\text{gen}}^{\ast})$  
      \State $\hat y\gets\text{softmax}(Wz+b)$                         
      \State $\mathcal L\gets\text{CE}(y,\hat y)$
    \EndFor
    \State update $\Theta$ 
  \EndFor
\EndFor
\end{algorithmic}
\end{algorithm}

\begin{algorithm}
\caption{TriShGAN}
\label{alg:trishgan}
\begin{algorithmic}[1]
\Require Training dataset $\mathcal{D}_{\text{train}} = \{\mathcal{X}_i, \mathbf{y}_i)\}_{i=1}^{N}$, Pre-trained Shapelet Extractor $\mathcal E_{\Theta}$, Pretrained classifier $C$, Number of epochs $T$, Loss weights $\lambda_1, \lambda_2, \lambda_3, \lambda_4$
\Ensure Counterfactual explanation $\mathcal{X}_{cf}$ for queried time series $\mathcal{X}_q$

\State \textbf{Step 1: Extract Discriminative Subsequence}
\State $I \gets$ %
      \Call{Shapelet Extractor}{$\mathcal{D}_{\text{train}}$}

\State \textbf{Step 2: Train CounteRGAN (Generator $G$ and Discriminator $D$)}
\For{$t = 1$ to $T$}
    \For{each batch $(\mathcal{X}, \mathbf{y})$ from $\mathcal{D}_{\text{train}}$}
        \State Extract shapelet subsequence set: $I$
        \State $\Delta \mathcal{X} \leftarrow G(I)$ \Comment {Generate residual perturbation}
        \State $ \mathcal{X_\text{cf}} =  \mathcal{X} + \Delta  \mathcal{X}$ \Comment {Compute counterfactual}
    
        \State $L_{RGAN} = \mathbb{E}[\log D(\mathcal{X})] + \mathbb{E}[\log(1 - D({\mathcal{X}+\Delta \mathcal{X}}))]$ \Comment{Adversarial loss}

        \State $L_G = \lambda_1 L_{triplet} + \lambda_2 L_{RGAN} + \lambda_3 L_{C} + \lambda_4 L_{0} + \lambda_5 L_{1}$ 
        \State Update generator: $G \gets G - \eta \nabla_G L_G$
        \State Update discriminator: $D \gets D - \eta \nabla_D L_{RGAN}$
    \EndFor
\EndFor

\State \textbf{Step 3: Generate Counterfactual Explanation}
\State Extract shapelet representation: $I \leftarrow \mathcal{E}(\mathcal{X_\text{test}})$
\State Generate residual perturbation: $\Delta \mathcal{X\text{test}} \leftarrow G(\mathcal{X\text{test}})$
\State Compute final counterfactual: ${\mathcal{X}_{cf}} = \mathcal{X_\text{test}} + \Delta \mathcal{X_\text{test}}$

\State \Return $\mathcal{X}_{cf}$
\end{algorithmic}
\end{algorithm}

\section{Baseline and Metrics}
\subsection{Baseline}
We evaluate our method against other baseline methods, including a NUN-based method AB-CF \cite{li2023attention}, and some GAN-based explanation methods such as GAN \cite{goodfellow2020generative}, CounteRGAN \cite{nemirovsky2022countergan} and SPARSE \cite{lang2023generating}. 
\begin{itemize}
    \item \textbf{Attention-based Counterfactual Explanation (AB-CF):} This method utilizes Shannon entropy to identify the most important $k$ subsequences within a queried time series. It quantifies the information contained in each segment based on its probability distribution and then uses \cite{delaney2021instance} paradigm to generate the CFE.
    \item \textbf{GAN:} This method produces the complete CFE rather than just residuals. The generator loss is regularized using the $\mathcal{L}_1$ and $\mathcal{L}_0$ norm to enhance the similarity and sparsity of the counterfactual. The fully-connected output layer of the generator is followed by a Tanh activation function.
    
    \item \textbf{CounteRGAN (C-GAN)} This method produces a residual based on the queried time series by calculating the subtraction result from two fully connected layers. Other aspects of the implementation are the same as the GAN approach.

    \item \textbf{SPAESE:} This method is similar to CounteRGAN, but utilizes a custom sparsity layer consisting of two interoperating ReLU activations. The generator employs a bi-LSTM for forward/backward dependencies, then passes its hidden state through two fully connected layers (each followed by ReLU). One outputs a nonnegative positive shift, and the other a nonnegative negative shift. The residual is computed as: $\text{residual} = \text{RELU}(fc1(\text{hidden})) - \text{RELU}(fc2(\text{hidden}))$.
\end{itemize}

\subsection{Metrics}
\subsubsection{Local Outlier Factor (LOF)}
We provide further details on the LOF. LOF quantifies how much a $X^{cf}$ deviates from the norm within a baseline dataset \(\mathcal{X}_{\text{test}}\) \cite{patcha2007overview}. A LOF greater than 1 indicates that \(X^{cf}\) is significantly less dense than its neighbors, classifying it as an outlier. It is calculated as: \begin{align} LOF_{k,s}(X^{cf}) = \frac{\sum_{{X^{cf}}' \in L_k(X^{cf})} \left(\frac{lrd_k(\mathbf{x}')}{lrd_k(X^{cf})}\right)}{|L_k(X^{cf})|}, \end{align} where \( L_k(X^{cf}) \) represents the k-nearest neighbors of \(X^{cf}\) in \(\mathcal{X}_{\text{test}}\). 
\subsubsection{Robustness}
For binary classification, we define robustness as the distance of a counterfactual instance from the decision boundary.  
Formally, $R\!\bigl(X_{\mathrm{cf}}\bigr)\;=\;\bigl|\,P\!\bigl(X_{\mathrm{cf}}\bigr)-0.5\bigr|,$
where $P(X_{\mathrm{cf}})$ denotes the model’s predicted probability of the positive class. Larger values indicate that small perturbations are unlikely to revert the counterfactual to its original label. To verify the stability of this metric, we injected Gaussian noise with varying standard deviations into the input before passing it to the generator $G$. Using the FM dataset, we found that the robustness score remained high across all noise levels (Table \ref{tab:gauss_noise}).

\begin{table}[h]
\centering
\caption{Robustness under gaussian noise with different noise scales.}
\label{tab:gauss_noise}
\begin{tabular}{@{}lcccc@{}}
\toprule
Scale            & 0        & 0.2        & 0.4        & 0.6 \\ \midrule
Robustness & 0.009\,$\pm$\,0.002 & 0.007\,$\pm$\,0.001 & 0.011\,$\pm$\,0.004 & 0.012\,$\pm$\,0.003 \\ \bottomrule
\end{tabular}
\end{table}

\section{Dataset}
We use  four publicly available MTS real-world datasets from the UEA MTS archive: FingerMovements (FM), HeartBeat (HB), SelfRegulationSCP1 (SCP1), and SelfRegulationSCP2 (SCP2). Details is showed in Table \ref{dataset}.
\begin{table}[htbp]
\centering
\caption{Dataset}
\label{dataset}
\small
\setlength{\tabcolsep}{1pt} 
\begin{tabular}{lcccccc}
\toprule
Dataset              & \makecell[c]{Window\\Size}
 & \makecell[c]{Number of\\ Shapelets}  & \makecell[c]{Number of\\Dimensions} & \makecell[c]{Length of\\Time Series} & \makecell[c]{Training\\Data} & \makecell[c]{Testing\\Data} \\
\midrule
FM     & 20          & 30                  & 28                   & 50                    & 316             & 100             \\
HB          & 200         & 100                 & 61                   & 405                    & 204             & 205             \\
SCP1   & 100         & 100                 & 6                    & 896                    & 268             & 293             \\
SCP2  & 100         & 100                 & 7                    & 1152                    & 200             & 180             \\
\bottomrule
\end{tabular}
\end{table}

\section{Complete Experiment Result}
We present a concise version of the main results in our paper while providing more detailed metrics (to three decimal places) in this section.
Table \ref{exp_results2}, \ref{ablation_shapelet2} and \ref{ablation_triplet2} showed the complete experiment result. For the plausibility metric, all methods exhibit good performance. First, for the AB-CF method, directly substituting a subsequence does not produce out-of-distribution counterfactuals. For GAN-based methods, the reference time series $X_{\text{reference}}$ comes from $\mathcal{X}_{\text{reference}}$ and is provided as input to the discriminator, with the model's objective being to generate counterfactual samples that closely resemble the distribution of the reference instance to deceive the discriminator. This mechanism promotes the generation of counterfactual samples that are more distributionally coherent.
\begin{table}[htbp]
\centering
\caption{Experiment Results Across Datasets and Methods}
\label{exp_results2}
\small
\begin{tabular}{lcccccc} 
\toprule
\multirow{2}{*}{Dataset} & \multirow{2}{*}{Method} & \multicolumn{5}{c}{Metrics} \\ 
\cmidrule(l){3-7}
 & & TCV $\uparrow$ & Robustness $\downarrow$ & Proximity $\downarrow$ & Sparsity $\downarrow$ & Plausibility $\downarrow$\\ 
\midrule
\multirow{5}{*}{FM} & AB-CF & $100.000 \pm 0.000$ & $0.306 \pm 0.035$ & $0.125 \pm 0.025$ & $0.460 \pm 0.096$ & $0.047 \pm 0.025$  \\


& GAN & $100.000 \pm 0.910$ & $0.040 \pm 0.011$ & $0.379 \pm 0.009$ & $1.00 0\pm 0.000$ & $1.000 \pm 0.000$\\
 & C-GAN & $97.959 \pm 2.041$ & $0.102 \pm 0.033$ & $0.752 \pm 0.545$ & $1.000 \pm 0.000$ & $0.035 \pm 0.002$\\
& SPARSE & $100.000 \pm 0.000$ & $0.035 \pm 0.009$ & $0.080 \pm 0.007$ & $0.108 \pm 0.003$ & $0.398 \pm 0.014$\\

 & Ours & $100.000 \pm 0.000$ & $0.012 \pm 0.009$ & $0.088 \pm 0.022$ & $0.036 \pm 0.008$ & $0.041 \pm 0.000$ \\

 \midrule
\multirow{5}{*}{HB} & AB-CF & $100.000 \pm 0.000$ & $0.336 \pm 0.010$ & $0.067 \pm 0.010$ & $0.365 \pm 0.070$ & $0.000 \pm 0.000$ \\


  & GAN & $100.000 \pm 0.000$ & $0.011 \pm 0.003$ & $0.171 \pm 0.003$ & $1.000 \pm 0.000$ & $0.000 \pm 0.000$ \\
  & C-GAN & $100.000 \pm 0.000$ & $0.134 \pm 0.033$ & $0.046 \pm 0.011$ & $1.000 \pm 0.000$ & $0.047 \pm 0.027$  \\
 & SPARSE & $100.000 \pm 0.000$ & $0.004 \pm 0.001$ & $0.023 \pm 0.006$ & $0.010 \pm 0.003$ & $0.421 \pm 0.000$\\

 & Ours & $100.000 \pm 0.000$ & $0.006 \pm 0.003$& $0.016 \pm 0.003$ & $0.006 \pm 0.003$ & $0.509 \pm 0.000$ \\

\midrule
\multirow{5}{*}{SCP1} & AB-CF & $100.000 \pm 0.000$ & $0.408 \pm 0.039$ & $0.449 \pm 0.142$ & $0.721 \pm 0.175$ & $0.115 \pm 0.130$ \\


& GAN & $100.000 \pm 0.000$ & $0.077 \pm 0.005$ & $0.373 \pm 0.001$ & $1.000 \pm 0.000$ & $0.000 \pm 0.000$\\
& C-GAN &  $98.402 \pm 2.202$ & $0.073 \pm 0.015$ & $0.068 \pm 0.019$ & $1.000 \pm 0.000$ & $0.049 \pm 0.027$\\
 & SPARSE & $98.630 \pm 1.182 $ & $0.038 \pm 0.004$ & $0.030 \pm 0.009$ & $0.009 \pm 0.004$ & $0.009 \pm 0.004$\\

 & Ours & $100.000 \pm 0.000$ & $0.003 \pm 0.001$ & $0.038 \pm 0.008$ & $0.012  \pm 0.002$ & $0.007  \pm 0.000$ \\

 \midrule
\multirow{5}{*}{SCP2} & AB-CF & $100.000 \pm 0.000$ & $0.396 \pm 0.026$ & $0.223 \pm 0.072$ & $0.435 \pm 0.133$ & $0.083 \pm 0.089$ \\


 & GAN & $100.000 \pm 0.000$ & $0.221 \pm 0.097$ & $0.247 \pm 0.010$ & $1.000 \pm 0.000$ & $0.000 \pm 0.000$ \\
  & C-GAN & $100.000 \pm 0.000$ & $0.171 \pm 0.007$ & $0.036 \pm 0.007$ & $1.000 \pm 0.000$ & $0.000 \pm 0.000$\\
 & SPARSE & $100.000 \pm 0.000 $ & $0.132 \pm 0.062$ & $0.030 \pm 0.011$ & $0.008 \pm 0.011$ & $0.011 \pm 0.000$\\

 & Ours & $100.000 \pm 0.000$ & $0.020  \pm 0.005$ & $0.031 \pm 0.002$ & $0.008  \pm 0.002$ & $0.011  \pm 0.000$
 \\

\bottomrule
\end{tabular}
\end{table}

\begin{table*}[htbp]
\centering
\caption{Ablation experiment for Shapelet Extractor}
\label{ablation_shapelet2}
\small
\begin{tabular}{lcccccc} 
\toprule
\multirow{2}{*}{Dataset} & \multirow{2}{*}{Method} & \multicolumn{5}{c}{Metrics} \\ 
\cmidrule(l){3-7}
 & & TCV $\uparrow$ & Robustness $\downarrow$ & Proximity $\downarrow$ & Sparsity $\downarrow$ & Plausibility $\downarrow$  \\ 
\midrule
\multirow{2}{*}{FingerMovements}
& without &  $100.000 \pm 0.000$ & $0.035 \pm 0.009$ & $0.080 \pm 0.007$ & $0.108 \pm 0.003$ & $0.398 \pm 0.014$\\

& with &  $100.000 \pm 0.000$ & $0.021 \pm 0.001$ & $0.094 \pm 0.018$ & $0.050 \pm 0.008$ & $0.048 \pm 0.012$\\

 \midrule
\multirow{2}{*}{Heartbeat} & without & $100.000 \pm 0.000$ & $0.004 \pm 0.001$ & $0.023 \pm 0.006$ & $0.010 \pm 0.003$ & $0.421 \pm 0.000$ \\

&  with &  $100.000 \pm 0.000$ & $0.003 \pm 0.001$ & $0.016 \pm 0.004$ & $0.006 \pm 0.002$ & $0.421 \pm 0.000$\\

\midrule
\multirow{2}{*}{SelfRegulationSCP1} & without & $98.630 \pm 1.182 $ & $0.038 \pm 0.004$ & $0.030 \pm 0.009$ & $0.009 \pm 0.004$ & $0.009 \pm 0.004$\\

&  with &  $100.00 \pm 0.000$ & $0.023 \pm 0.006$ & $0.043 \pm 0.024$ & $0.010\pm 0.006$ & $0.007 \pm 0.000$\\

\midrule
\multirow{2}{*}{SelfRegulationSCP2} & without & $100.00 \pm 0.00 $ & $0.132 \pm 0.062$ & $0.030 \pm 0.011$ & $0.008 \pm 0.011$ & $0.011 \pm 0.000$  \\

&  with &  $100.00 \pm 0.00$ & $0.067 \pm 0.052$ & $0.034 \pm 0.016$ & $0.006 \pm 0.003$ & $0.011 \pm 0.000$ \\

\bottomrule
\end{tabular}
\end{table*}

\begin{table*}[htbp]
\centering
\caption{Ablation experiment for triplet loss}
\small
\label{ablation_triplet2}
\begin{tabular}{lcccccc} 
\toprule
\multirow{2}{*}{Dataset} & \multirow{2}{*}{Method} & \multicolumn{5}{c}{Metrics} \\ 
\cmidrule(l){3-7}
 & & TCV $\uparrow$ & Robustness $\downarrow$ & Proximity $\downarrow$ & Sparsity $\downarrow$ & Plausibility $\downarrow$  \\ 
\midrule
\multirow{2}{*}{FingerMovements}
& without &  $54.433 \pm 5.356$ & $0.214 \pm 0.017$ & $0.089 \pm 0.010$ & $0.038 \pm 0.004$ & $0.214 \pm 0.017$\\

& with &  $97.959 \pm 0.000$  & $0.048 \pm 0.033$ & $0.102 \pm 0.029$ & $0.036 \pm 0.006$ & $1.000 \pm 0.000$ \\

 \midrule
\multirow{2}{*}{Heartbeat} & without & $99.415\pm 0.827$ & $0.066 \pm 0.047$ & $0.010 \pm 0.002$ & $0.016 \pm 0.017$ & $0.418 \pm 0.005$ \\

&  with &  $100.000 \pm 0.000$ & $0.004 \pm 0.001$  & $0.022 \pm 0.005$ & $0.007 \pm 0.002$ & $0.425 \pm 0.008$\\

\midrule
\multirow{2}{*}{SelfRegulationSCP1} & without & $26.712 \pm 0.000$ & $0.292 \pm 0.005$ & $0.001 \pm 0.000$ & $0.001 \pm 0.000$ & $0.026 \pm 0.000$\\

&  with &  100.000 ± 0.000 & 0.014 ± 0.009 & 0.019 ± 0.004 & 0.011 ± 0.005 & 0.007 ± 0.000 \\

\midrule
\multirow{2}{*}{SelfRegulationSCP2} & without & $44.815 \pm 2.313 $ & $0.435 \pm 0.006$ & $0.018 \pm 0.0018$ & $0.003 \pm 0.001$ & $0.017 \pm 0.015$  \\

&  with &  $100.000 \pm 0.000$ & $0.010 \pm 0.006$ & $0.003 \pm 0.001$ & $0.011 \pm 0.000$ & $0.188 \pm 0.021$ \\

\bottomrule
\end{tabular}
\end{table*}

\section{Compute Resources and Environment Details}
All experiments are run in a 16GB NVIDIA RTX A4000 GPU, the runtime per batch increases from 6.47 seconds to 6.97 seconds as grows from 2 to 8. Although a larger slightly raises training costs compared to not using triplet loss, we observe that TCV and robustness improve by approximately 80\%, and plausibility by about 52\%, thereby significantly enhancing the quality of CFEs.

\section{Further Analysis of Shaplet Extractor}

\subsubsection{Converge}
\begin{figure}[htbp]
\centering
\includegraphics[scale=0.12]{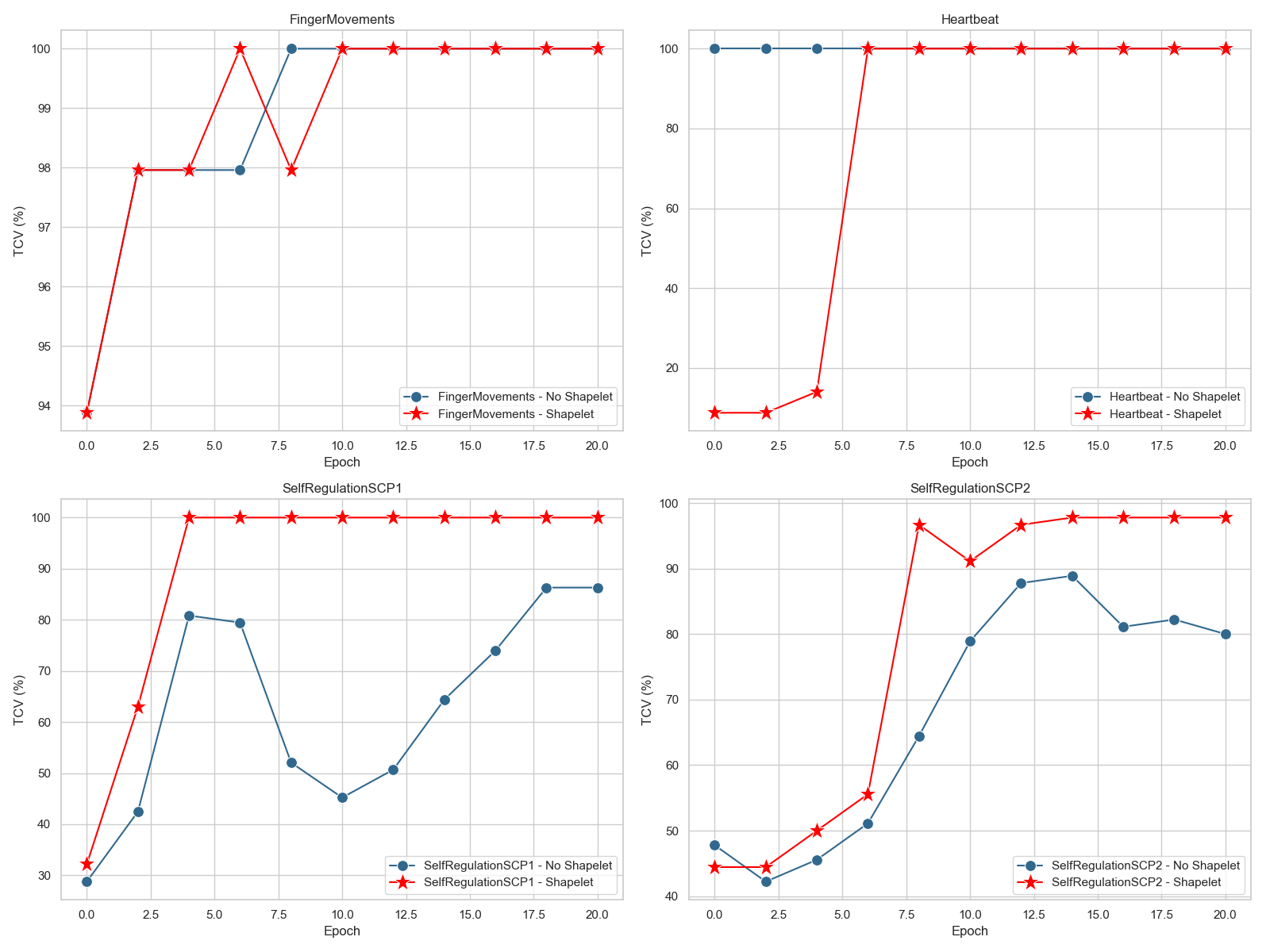}
\caption{Converge of TCV} 
\label{covergae_shapelet}
\end{figure}

Introducing a Shapelet Extractor into our framework enhances its ability to pinpoint and localize the discriminative parts within a time series. This precise localization facilitates a highly effective TCV. Figure \ref{covergae_shapelet} illustrates how the addition of shapelets can lead to a TCV approaching 100\% before the 10th epoch across all four datasets. 

\section{Hyperparameter Selection for Triplet loss}
We report the experimental results of the hyperparameter selection for the triplet loss in Table \ref{self1_triplet}, \ref{self2}, \ref{hearbeat} and \ref{Fingermovement},.

\begin{table*}[htbp]
\centering
\small
\caption{Triplet loss - SCP1}
\label{self1_triplet}
\begin{tabular}{cccccc}
\toprule
Margin & Validity  & Proximity & Sparsity & Plausibility  & Probability\\
\midrule
\multicolumn{6}{c}{$n$=2} \\
\midrule
600 & 100.000 ± 0.000 & 0.018 ± 0.011 & 0.029 ± 0.017 & 0.014 ± 0.006 & 0.007 ± 0.000 \\
700 & 100.000 ± 0.000 & 0.021 ± 0.013 & 0.020 ± 0.008 & 0.010 ± 0.006 & 0.007 ± 0.000 \\
800 & 68.493 ± 50.476 & 0.084 ± 0.059 & 0.012 ± 0.000 & 0.006 ± 0.001 & 0.005 ± 0.004 \\
900 & 98.858 ± 0.791 & 0.053 ± 0.024 & 0.011 ± 0.001 & 0.008 ± 0.002 & 0.009 ± 0.004 \\
\midrule
\multicolumn{6}{c}{$n$=4} \\
\midrule
600 & 100.000 ± 0.000 & 0.017 ± 0.009 & 0.021 ± 0.003 & 0.012 ± 0.007 & 0.007 ± 0.000 \\
700 & 100.000 ± 0.000 & 0.014 ± 0.009 & 0.019 ± 0.004 & 0.011 ± 0.005 & 0.007 ± 0.000 \\
800 & 98.630 ± 1.812 & 0.045 ± 0.026 & 0.016 ± 0.002 & 0.015 ± 0.002 & 0.007 ± 0.000 \\
900 & 100.000 ± 0.000 & 0.022 ± 0.014 & 0.018 ± 0.005 & 0.021 ± 0.005 & 0.009 ± 0.004 \\
\midrule
\multicolumn{6}{c}{$n$=6} \\
\midrule
600 & 100.000 ± 0.000 & 0.014 ± 0.010 & 0.037 ± 0.019 & 0.015 ± 0.004 & 0.007 ± 0.000 \\
700 & 100.000 ± 0.000 & 0.019 ± 0.002 & 0.034 ± 0.016 & 0.018 ± 0.004 & 0.007 ± 0.000 \\
800 & 100.000 ± 0.000 & 0.017 ± 0.014 & 0.043 ± 0.037 & 0.022 ± 0.020 & 0.007 ± 0.000 \\
900 & 95.434 ± 7.909 & 0.045 ± 0.023 & 0.012 ± 0.000 & 0.014 ± 0.005 & 0.007 ± 0.001 \\
\midrule
\multicolumn{6}{c}{$n$=8} \\
\midrule
600 & 100.000 ± 0.000 & 0.023 ± 0.017 & 0.074 ± 0.089 & 0.028 ± 0.020 & 0.023 ± 0.028 \\
700 & 99.543 ± 0.791 & 0.043 ± 0.019 & 0.017 ± 0.006 & 0.014 ± 0.005 & 0.007 ± 0.000 \\
800 & 100.000 ± 0.000 & 0.008 ± 0.010 & 0.029 ± 0.013 & 0.014 ± 0.006 & 0.009 ± 0.004 \\
900 & 99.772 ± 0.395 & 0.020 ± 0.001 & 0.014 ± 0.001 & 0.020 ± 0.005 & 0.007 ± 0.000 \\

\bottomrule
\end{tabular}
\end{table*}

\begin{table*}[htbp]
\centering
\small
\caption{Triplet loss - SCP2}
\label{self2}
\begin{tabular}{cccccc}
\toprule
Margin & Validity & Proximity & Sparsity & Plausibility & Probability\\
\midrule
\multicolumn{6}{c}{$n$ = 2} \\
\midrule
600 & $99.630 \pm 0.642$ & $0.030 \pm 0.004$ & $0.006 \pm 0.003$ & $0.007 \pm 0.006$ & $0.124 \pm 0.016$ \\
700 & $99.630 \pm 0.642$ & $0.026 \pm 0.026$ & $0.023 \pm 0.035$ & $0.011 \pm 0.000$ & $0.133 \pm 0.055$ \\
800 & $99.630 \pm 0.642$ & $0.011 \pm 0.010$ & $0.003 \pm 0.002$ & $0.011 \pm 0.000$ & $0.212 \pm 0.064$ \\
900 & $100.000 \pm 0.000$ & $0.010 \pm 0.006$ & $0.003 \pm 0.001$ & $0.011 \pm 0.000$ & $0.188 \pm 0.021$ \\
\midrule
\multicolumn{6}{c}{$n$ = 4} \\
\midrule
600 & $100.000 \pm 0.000$ & $0.277 \pm 0.359$ & $0.107 \pm 0.161$ & $0.007 \pm 0.006$ & $0.058 \pm 0.008$ \\
700 & $100.000 \pm 0.000$ & $0.040 \pm 0.014$ & $0.012 \pm 0.002$ & $0.011 \pm 0.000$ & $0.078 \pm 0.037$ \\
800 & $100.000 \pm 0.000$ & $0.027 \pm 0.022$ & $0.007 \pm 0.001$ & $0.011 \pm 0.000$ & $0.152 \pm 0.098$ \\
900 & $99.630 \pm 0.642$ & $0.014 \pm 0.007$ & $0.004 \pm 0.002$ & $0.011 \pm 0.000$ & $0.137 \pm 0.032$ \\

\midrule
\multicolumn{6}{c}{$n$ = 6} \\
\midrule
600 & $100.000 \pm 0.000$ & $1.298 \pm 0.519$ & $0.503 \pm 0.071$ & $0.007 \pm 0.006$ & $0.098 \pm 0.022$ \\
700 & $100.000 \pm 0.000$ & $0.181 \pm 0.092$ & $0.075 \pm 0.054$ & $0.030 \pm 0.032$ & $0.080 \pm 0.053$ \\
800 & $100.000 \pm 0.000$ & $0.032 \pm 0.014$ & $0.008 \pm 0.003$ & $0.011 \pm 0.000$ & $0.085 \pm 0.043$ \\
900 & $99.630 \pm 0.642$ & $0.030 \pm 0.012$ & $0.008 \pm 0.005$ & $0.011 \pm 0.000$ & $0.102 \pm 0.054$ \\
\midrule
\multicolumn{6}{c}{$n$ = 8} \\
\midrule
600 & $100.000 \pm 0.000$ & $1.280 \pm 0.494$ & $0.292 \pm 0.006$ & $0.015 \pm 0.006$ & $0.139 \pm 0.031$ \\
700 & $99.630 \pm 0.642$ & $0.705 \pm 0.608$ & $0.322 \pm 0.264$ & $0.019 \pm 0.023$ & $0.116 \pm 0.072$ \\
800 & $99.259 \pm 1.283$ & $0.054 \pm 0.028$ & $0.014 \pm 0.006$ & $0.011 \pm 0.000$ & $0.126 \pm 0.010$ \\
900 & $100.000 \pm 0.000$ & $0.047 \pm 0.033$ & $0.006 \pm 0.003$ & $0.011 \pm 0.000$ & $0.048 \pm 0.014$ \\

\bottomrule
\end{tabular}
\end{table*}

\begin{table*}[htbp]
\centering
\small
\caption{Triplet loss - HB}
\label{hearbeat}
\begin{tabular}{cccccc}
\toprule
Margin & Validity & Proximity & Sparsity & Plausibility & Robustness  \\
\midrule
\multicolumn{6}{c}{$n$ = 2} \\
\midrule
50 & $100.000 \pm 0.000$ & $0.022 \pm 0.005$ & $0.007 \pm 0.002$ & $0.425 \pm 0.008$ & $0.004 \pm 0.001$  \\
 60 & $99.649 \pm 0.785$ & $0.019 \pm 0.009$ & $0.007 \pm 0.003$ & $0.419 \pm 0.005$ & $0.011 \pm 0.013$\\
70 & $100.000 \pm 0.000$ & $0.017 \pm 0.004$ & $0.009 \pm 0.006$ & $0.421 \pm 0.000$ & $0.027 \pm 0.017$ \\
80 & $81.579 \pm 33.364$ & $0.011 \pm 0.001$ & $0.010 \pm 0.004$ & $0.303 \pm 0.203$ & $0.116 \pm 0.133$  \\
\midrule
\multicolumn{6}{c}{$n$ = 4} \\
\midrule
50 & $94.737 \pm 11.769$ & $0.015 \pm 0.003$ & $0.011 \pm 0.013$ & $0.389 \pm 0.071$ & $0.049 \pm 0.097$\\
60 & $98.684 \pm 0.877$ & $0.016 \pm 0.005$ & $0.011 \pm 0.004$ & $0.422 \pm 0.017$ & $0.020 \pm 0.020$ \\
70 & $49.123 \pm 27.292$ & $0.166 \pm 0.002$ & $0.054 \pm 0.004$ & $0.103 \pm 0.145$ & $0.328 \pm 0.098$ \\
80 & $36.842 \pm 3.745$ & $0.474 \pm 0.032$ & $0.100 \pm 0.001$ & $0.000 \pm 0.001$ & $0.297 \pm 0.02$ \\
\midrule
\multicolumn{6}{c}{$n$ = 6} \\
\midrule
50 & $100.000 \pm 0.000$ & $0.019 \pm 0.004$ & $0.007 \pm 0.003$ & $0.432 \pm 0.024$ & $0.038 \pm 0.063$ \\
60 & $97.368 \pm 3.359$ & $0.195 \pm 0.340$ & $0.043 \pm 0.035$ & $0.410 \pm 0.013$ & $0.155 \pm 0.098$  \\
70 & $99.123 \pm 1.241$ & $0.170 \pm 0.020$ & $0.074 \pm 0.001$ & $0.416 \pm 0.007$ & $0.020 \pm 0.003$\\
80 & $61.404 \pm 54.584$ & $0.291 \pm 0.084$ & $0.080 \pm 0.008$ & $0.211 \pm 0.298$ & $0.183 \pm 0.209$\\
\midrule
\multicolumn{6}{c}{$n$ = 8} \\
\midrule
50 & $99.649 \pm 0.785$ & $0.015 \pm 0.002$ & $0.018 \pm 0.023$ & $0.426 \pm 0.011$ & $0.011 \pm 0.009$\\
60 & $78.947 \pm 21.391$ & $0.103 \pm 0.036$ & $0.064 \pm 0.014$ & $0.287 \pm 0.131$ & $0.127 \pm 0.115$  \\
70 & $96.491 \pm 0.000$ & $0.156 \pm 0.017$ & $0.072 \pm 0.004$ & $0.391 \pm 0.013$ & $0.058 \pm 0.029$\\
80 & $87.719 \pm 0.000$ & $0.268 \pm 0.030$ & $0.090 \pm 0.000$ & $0.340 \pm 011$ & $0.152 \pm 0.008$ \\

\bottomrule
\end{tabular}
\end{table*}

\begin{table*}[htbp]
\centering
\small
\caption{Triplet loss - FM}
\label{Fingermovement}
\begin{tabular}{ccccccc}
\toprule
Margin & Validity & Proximity & Sparsity & Plausibility & Robustness\\
\midrule
\multicolumn{6}{c}{$n$ = 2} \\
\midrule
185 & $90.476 \pm 12.961$ & $0.084 \pm 0.015$ & $0.033 \pm 0.011$ & $1.000 \pm 0.000$ & $0.058 \pm 0.062$ \\
285 & $70.408 \pm 33.191$ & $0.085 \pm 0.006$ & $0.036 \pm 0.013$ & $1.000 \pm 0.000$ & $0.239 \pm 0.217$\\
385 & $80.612 \pm 27.418$ & $0.068 \pm 0.026$ & $0.027 \pm 0.011$ & $1.000 \pm 0.000$ & $0.146 \pm 0.160$  \\
485 & $71.429 \pm 31.879$ & $0.089 \pm 0.020$ & $0.193 \pm 0.058$ & $1.000 \pm 0.000$ & $0.144 \pm 0.004$ \\

\midrule
\multicolumn{6}{c}{$n$ = 4} \\
\midrule
185 & $69.388 \pm 35.874$ & $0.098 \pm 0.002$ & $0.035 \pm 0.004$ & $0.667 \pm 0.577$ & $0.222 \pm 0.106$  \\
285 & $67.347 \pm 21.598$ & $0.050 \pm 0.022$ & $0.022 \pm 0.003$ & $1.000 \pm 0.000$ & $0.132 \pm 0.079$ \\
385 & $72.109 \pm 36.297$ & $0.076 \pm 0.041$ & $0.127 \pm 0.122$ & $1.000 \pm 0.000$ & $0.150 \pm 0.088$ \\
485 & $57.143 \pm 23.089$ & $0.240 \pm 0.053$ & $0.516 \pm 0.109$ & $1.000 \pm 0.000$ & $0.270 \pm 0.149$ \\

\midrule
\multicolumn{6}{c}{$n$ = 6} \\
\midrule
185 & $97.959 \pm 0.000$ & $0.102 \pm 0.029$ & $0.036 \pm 0.006$ & $1.000 \pm 0.000$ & $0.048 \pm 0.033$ \\
285 & $93.878 \pm 4.082$ & $0.088 \pm 0.044$ & $0.040 \pm 0.017$ & $1.000 \pm 0.000$ & $0.053 \pm 0.057$\\
385 & $62.245 \pm 41.849$ & $0.060 \pm 0.018$ & $0.044 \pm 0.010$ & $1.000 \pm 0.000$ & $0.137 \pm 0.003$ \\
485 & $59.184 \pm 14.431$ & $0.403 \pm 0.008$ & $0.670 \pm 0.031$ & $1.000 \pm 0.000$ & $0.293 \pm 0.145$ \\

\midrule
\multicolumn{6}{c}{$n$ = 8} \\
\midrule
185 & $81.633 \pm 21.307$ & $0.095 \pm 0.013$ & $0.035 \pm 0.009$ & $1.000 \pm 0.000$ & $0.127 \pm 0.095$ \\
285 & $92.517 \pm 1.178$ & $0.097 \pm 0.029$ & $0.037 \pm 0.013$ & $1.000 \pm 0.000$ & $0.084 \pm 0.071$\\
385 & $91.156 \pm 8.248$ & $0.060 \pm 0.009$ & $0.027 \pm 0.003$ & $1.000 \pm 0.000$ & $0.108 \pm 0.090$\\
485 & $61.224 \pm 46.178$ & $0.468 \pm 0.013$ & $0.712 \pm 0.014$ & $1.000 \pm 0.000$ & $0.250 \pm 0.159$ \\
\bottomrule
\end{tabular}
\end{table*}

\section{Some Discussions}
\begin{itemize}
    \item We clarify the the green point does not mean the label has been flipped successfully. The discriminator distinguishes between real reference and generated CF samples, ensuring that the generated CF aligns with the distribution of the reference class, thus improving plausibility. The classifier ensures that the generated CF flips the label from the original class to the desired class. Marking a CF sample as real only indicates that it appears similar to the reference data distribution but does not guarantee crossing the classifier’s decision boundary. We also conducted an experiment without the classifier in our framework, and the results were poor: the TCV dropped to ~50\% for most datasets (except for HB).
    \item We clarify why change the discriminative subsequence of a queried time series. Changing a \emph{non‑discriminative} portion may indeed alter the prediction, but modifying such regions lacks interpretability and is costly. MTS typically span many signals and time steps, yet the decisive information is often concentrated in a few key subsequences. A \emph{shapelet extractor} therefore locates the subsequence~$s$ with the highest information gain~(IG), which measures how well a distance–based split separates the classes. Specifically, given an optimal threshold~$\delta^{\ast}$, the training set~$D$ is partitioned as $
  D_{1} = \{\,X \in D \mid \operatorname{dist}(X,s) \le \delta^{\ast}\,\},
  \qquad
  D_{2} = D \setminus D_{1},
$
and the information gain is
$
  \operatorname{IG}(s,D)
    = H(D)
      \;-\;
      \sum_{i=1}^{2}
      \frac{|D_{i}|}{|D|}\,H(D_{i}),
$
where $H(\cdot)$ denotes the entropy over class labels. Because the chosen shapelet~$s^{\ast}$ maximises this IG, modifying only the corresponding high‑IG subsequence in a test series~$X$ so that $\operatorname{dist}(X,s^{\ast})$ crosses~$\delta^{\ast}$ is typically the most efficient and \emph{sparse} way to flip the prediction. In contrast, adjusting low‑IG (non‑discriminative) regions generally requires disproportionately large perturbations and is therefore less cost‑effective for achieving the desired outcome.

\end{itemize}

\end{document}